\documentclass[graybox,envcountsame]{svmult}

\usepackage{mathptmx}       
\usepackage{helvet}         
\usepackage{courier}        
\usepackage{type1cm}        
\usepackage{makeidx}         
\usepackage{graphicx}        
\usepackage{multicol}        
\usepackage[bottom]{footmisc}

\usepackage{amsmath} 
\usepackage{amssymb}
\usepackage{mathtools}
\usepackage{bbm}
\usepackage{latexsym}
\usepackage{dsfont}
\usepackage{pifont}

\usepackage{color} 
\usepackage{xcolor} 
\usepackage{algorithm2e}
\usepackage{tikz}
\usetikzlibrary{positioning,shadings,arrows, decorations.markings}
\usepackage{pgfplots}

\usepackage{verbatim}

\usepackage{blkarray}
\usepackage{hyperref}
\definecolor{bl}{RGB}{20,20,150}
\usepackage{url}

\newcommand{\Ecal}{\mathcal{E}}

\newcommand{\Mcal}{\mathcal{M}}
\newcommand{\Pcal}{\mathcal{P}}

\newcommand{\Xcal}{\mathcal{X}}

\newcommand{\RBM}{\operatorname{RBM}}
\newcommand{\supp}{\operatorname{supp}}

\newcommand{\Ycal}{\mathcal{Y}}

\newcommand{\argmax}{\operatorname{argmax}}

\newcommand\independent{\protect\mathpalette{\protect\independenT}{\perp}}
\def\independenT#1#2{\mathrel{\rlap{$#1#2$}\mkern2mu{#1#2}}}

\usepackage{tcolorbox}
\makeatletter
\newcommand{\mybox}[1]{%
	\setbox0=\hbox{#1\!\!}%
	\setlength{\@tempdima}{\dimexpr\wd0+13pt}%
	\begin{tcolorbox}[colframe=gray, boxrule=1.5pt, arc=4pt,
		left=6pt,right=6pt, top=2pt,bottom=2pt,boxsep=0pt,width=\@tempdima]
		#1
	\end{tcolorbox}
}

\definecolor{bl}{RGB}{20,20,150}
\definecolor{gr}{RGB}{20,180,20}
\definecolor{ic}{RGB}{180,20,20}
\definecolor{oc}{RGB}{20,180,20}

\begin{document}
	
	\title*{Restricted Boltzmann Machines:\\ Introduction and Review}
	\titlerunning{RBMs: Introduction and Review}
	\author{Guido Mont\'ufar}
	\authorrunning{G.~Mont\'ufar} 
	\institute{Guido Mont\'ufar 
	\at Department of Mathematics and Department of Statistics, University of California, Los Angeles, USA. 
		\email{montufar@math.ucla.edu}
}

	\maketitle

\abstract{The restricted Boltzmann machine is a network of stochastic units with undirected interactions between pairs of visible and hidden units. 
This model was popularized as a building block of deep learning architectures and has continued to play an important role in applied and theoretical machine learning. 
Restricted Boltzmann machines carry a rich structure, with connections to geometry, applied algebra, probability, statistics, machine learning, and other areas. 
The analysis of these models is attractive in its own right and also as a platform to combine and generalize mathematical tools for graphical models with hidden variables. 
This article gives an introduction to  the mathematical analysis of restricted Boltzmann machines, reviews recent results on the geometry of the sets of probability distributions representable by these models, and suggests a few directions for further investigation. }


\section{Introduction}

This article is intended as an introduction to the mathematical analysis of the restricted Boltzmann machine. 
Complementary to other existing and excellent introductions, we emphasize mathematical structures in relation to the geometry of the set of distributions that can be represented by this model. 
%
There is a large number of works on theory and applications of restricted Boltzmann machines. 
We review a selection of recent results in a way that, so we hope, can serve as a guide to this rich subject, and lets us advertise some of the interesting and challenging problems that still remain to be addressed. 

\paragraph{Brief overview}

%
A Boltzmann machine is a model of pairwise interacting units that update their states over time in a probabilistic way depending on the states of the adjacent units. 
Boltzmann machines have been motivated as models for parallel distributed computing~\cite{hinton1983, Ackley85alearning,Hinton:1986:LRB:104279.104291}. 
They can be regarded as stochastic versions of Hopfield networks~\cite{Hopfield:1988:NNP:65669.104422}, which serve as associative memories. 
They are closely related to mathematical models of interacting particles studied in statistical physics, especially the Ising model~\cite[Chapter~14]{huang2000statistical}. 
For each fixed choice of interaction strengths and biases in the network, 
the collective of units assumes different states at relative frequencies that depend on their associated energy, in what is known as a Gibbs-Boltzmann probability distribution~\cite{gibbs1902elementary}. 
As pair interaction models, Boltzmann machines define special types of hierarchical log-linear models, 
which are special types of exponential family models~\cite{Brown86:Fundamentals_of_Exponential_Families} closely related to undirected graphical models~\cite{lauritzen1996,jordan2004}. 
In contrast to the standard discussion of exponential families, Boltzmann machines usually involve hidden variables. 
Hierarchical log-linear models are widely used in statistics. Their geometric properties are studied especially in information geometry~\cite{Amari99informationgeometry,amari2007methods,igaia,InfGeo} and algebraic statistics~\cite{drton2009lectures,sullivant2018}. 
The information geometry of the Boltzmann machine was first studied by Amari, Kurata, and Nagaoka~\cite{125867}. 

%
A restricted Boltzmann machine (RBM) is a special type of a Boltzmann machine where the pair interactions are restricted to be between an observed set of units and an unobserved set of units. 
These models were introduced in the context of harmony theory~\cite{Smolensky:1986:IPD:104279.104290} and unsupervised two layer networks~\cite{NIPS1991_535}. 
RBMs played a key role in the development of greedy layer-wise learning algorithms for deep layered architectures~\cite{Hinton:2006:FLA:1161603.1161605,Bengio-2009}. 
A recommended introduction to RBMs is~\cite{10.1007/978-3-642-33275-3_2}. 
RBMs have been studied intensively, with tools from optimization, algebraic geometry, combinatorics, coding theory, polyhedral geometry, and information geometry among others. 
Some of the advances over the past few years include results in relation to their approximation properties~\cite{Younes1996109,
	LeRoux:2008:RPR:1374176.1374187,
	NIPS2011_0307,
	montufar2016hierarchical}, 
dimension~\cite{Cueto2010,montufar2013discrete, montufar2015dimension}, semi-algebraic description~\cite{Cueto:2010:ICB:1866469.1866627, SeigalMontufar}, 
efficiency of representation~\cite{NIPS2013_5020, montufar2015does}, 
sequential optimization~\cite{fischer2010bounding, Fischer15PTBound},  
statistical complexity~\cite{aoyagi:2010}, 
sampling and training~\cite{Salakhutdinov08learningand,fischer2009contrastive,fischer2010bounding,Fischer15PTBound}, 
information geometry~\cite{125867,igaia,Karakida:2016:DAC:2949079.2949224}. 
%

\paragraph{Organization}

This article is organized as follows. 
In Section~\ref{sec:boltzmannmachines} we introduce Boltzmann machines, Gibbs sampling, and the associated probability models. 
In Section~\ref{sec:definitions} we introduce restricted Boltzmann machines and discuss various perspectives, viewing the probability models as marginals of exponential families with Kronecker factoring sufficient statistics, as products of mixtures of product distributions, and as feedforward networks with soft-plus activations. 
We also discuss a piecewise linear approximation called tropical RBM model, which corresponds to a feedforward network with rectified linear units. 
In Section~\ref{sec:training} we give a brief introduction to training by maximizing the likelihood of a given data set. 
We comment on gradient, contrastive divergence, natural gradient, and EM methods. 
Thereafter, in Section~\ref{sec:dimension} we discuss the Jacobian of the model parametrization and the model dimension. 
In Section~\ref{sec:representational} we discuss the representational power, covering two hierarchies of representable distributions, 
namely mixtures of product distributions and hierarchical log-linear models, depending on the number of hidden units of the RBM. 
In Section~\ref{sec:approximation} we use the representation results to obtain bounds on the approximation errors of RBMs. 
In Section~\ref{sec:semialgebraic} we discuss semi-algebraic descriptions and a recent result for a small RBM. 
Finally, in Section~\ref{sec:outlook} we collect a few open questions and possible research directions.

\section{Boltzmann machines}
\label{sec:boltzmannmachines}

A Boltzmann machine is a network of stochastic units. 
Each unit, or neuron, can take one of two states. 
The joint state of all units has an associated energy value which is determined by pair interactions and biases. 
%
%
The states of the units are updated in a stochastic manner at discrete time steps, 
whereby lower energy states are preferred over higher energy ones. 
In the limit of infinite time, the relative number of visits of each state, or the relative probability of observing each state, converges to a fixed value that is exponential in the energy differences. 
The set of probability distributions that result from all possible values of the pair interactions and biases, forms a manifold of probability distributions called Boltzmann machine probability model. 
The probability distributions for a subset of visible units are obtained via marginalization, adding the probabilities of all joint states that are compatible with the visible states. 
We make these notions more specific in the following.

\paragraph{Pairwise interacting units}

We consider a network defined by a finite set of nodes $N$ and a set of edges $I\subseteq {N\choose 2}$ connecting pairs of nodes. 
Each node $i\in N$ corresponds to a random variable, or unit, with states $x_i \in \{0,1\}$. 
The joint states of all units are vectors $x=(x_i)_{i\in N}\in\{0,1\}^N$. 
Each unit $i\in N$ has an associated bias $\theta_i\in\mathbb{R}$, and each edge $\{i,j\}\in I$ has an associated interaction weight $\theta_{\{i,j\}}\in\mathbb{R}$. 
For any given value of the parameter $\theta=((\theta_i)_{i\in N},(\theta_{\{i,j\}})_{\{i,j\}\in I})$, the energy of the joint states $x$ is given by 
\begin{equation}
E(x; \theta) = - \sum_{i \in N} \theta_i x_i - \sum_{ \{  i,j \}\in I} \theta_{\{ i,j\} } x_i x_j,  \quad x\in \{0,1\}^{N} . 
\label{eq:energybm}
\end{equation} 
In particular, the negative energy function $-E(\cdot;\theta)$ is a linear combination of the functions $x\mapsto x_i$, $i\in N$, $x\mapsto x_ix_j$, $\{i,j\}\in I$, with coefficients $\theta$. It takes lower values when pairs of units with positive interaction take the same states, or also when units with positive bias take state one.

\paragraph{State updates, Gibbs sampling}

The Boltzmann machine updates the states of its units at discrete time steps, in a process known as Gibbs sampling. 
Given a state $x^{(t)}\in\{0,1\}^N$ at time $t$, the state $x^{(t+1)}$ at the next time step is created by selecting a unit $i \in N$,  
and then setting $x^{(t+1)}_i=1$ with probability 
\begin{equation}
\Pr\big(x^{(t+1)}_i=1 | x^{(t)}\big) = \sigma\Big(\sum_{\{i,j\}\in I}\theta_{\{i,j\}}x_j^{(t)} + \theta_i \Big) , 
\label{eq:transitionprob}
\end{equation} 
or $x^{(t+1)}_i=0$ with complementary probability $\Pr(x^{(t+1)}_i=0 | x^{(t)}) = 1 - \Pr(x^{(t+1)}_i=1 | x^{(t)})$. 
Here $\sigma\colon s \mapsto 1/(1 + \exp(-s))$ is the standard logistic function. 
In particular, the quotient of the probabilities of setting either $x_i=1$ or $x_i=0$ is the exponential energy difference $\sum_{\{i,j\}\in I}\theta_{\{i,j\}}x_j + \theta_i$ between the two resulting joint states. 
The activation probability~\eqref{eq:transitionprob} can be regarded as the output value of a deterministic neuron with inputs $x_{j}$ weighted by $\theta_{\{i,j\}}$ for all adjacent $j$s, bias $\theta_i$,  and activation function $\sigma$. 

If the unit $i$ to be updated at time $t$ is selected according to a probability distribution $r$ over $N$, and $T_i(x^{(t+1)}|x^{(t)})$ denotes the Markov transition kernel when choosing unit $i$, then the altogether transition kernel is 
\begin{equation*}
T =  \sum_{i\in N } r(i) T_i. 
\end{equation*}
In other words, if the state at time $t$ is $x^{(t)}$, then the state $x^{(t+1)}$ at the next time step is drawn from the probability distribution $T(\cdot|x^{(t)})$. 
More generally, if $p^{(t)}$ is a probability distribution over joint states $x^{(t)}\in \{0,1\}^N$ at time $t$, 
then at time $t+1$ we have the probability distribution 
\begin{equation*}
p^{(t+1)} = p^{(t)} \cdot T.  
\end{equation*}
The one step transition kernel $T$ is non zero only between state vectors $x^{(t)}$ and $x^{(t+1)}$ that differ at most in one entry. 
However, if $r$ is strictly positive, then there is a positive probability of transitioning from any state to any other state in $N$ time steps, so that the $N$-th power $T^N$ is strictly positive, implying that $T$ is a primitive kernel.

\paragraph{Stationary limit distributions}

If $T$ is a primitive kernel, then there is a unique distribution $p$ with $\lim_{t\to\infty} p^0\cdot T^t = p$, for all start state distributions $p^0$. 
This follows from a theorem by Geman and Geman, which also shows that $p$ is the Gibbs-Boltzmann distribution 
\begin{equation}
	p(x;\theta)=\frac{1}{Z(\theta)} \exp(-E(x;\theta)),\quad x\in \{0,1\}^N, 
	\label{eq:limit}
\end{equation}
with the energy function $E(\cdot;\theta)$ given in~\eqref{eq:energybm} and normalizing partition function 
$Z(\theta)=\sum_{x'}\exp(-E(x';\theta))$. \\

The set of stationary distributions~\eqref{eq:limit}, for all $\theta\in\mathbb{R}^{|N|+ |I|}$, is the Boltzmann machine probability model with interaction structure $G=(N,I)$. 
This is an exponential family with sufficient statistics $x_i, i\in N$, $x_ix_j$, $\{i,j\}\in I$ and canonical or exponential parameter $\theta$. It is a smooth manifold of dimension $|N|+|I|$, contained in the $2^{N}-1$ dimensional simplex of probability distributions on $\{0,1\}^N$, 
\begin{equation*}
\Delta_{\{0,1\}^N} = \Big\{p\in\mathbb{R}^{\{0,1\}^N}\colon p(x)\geq 0\;\text{ for all }\; x\in\{0,1\}^N,\; \text{ and }\;  \sum_{x\in\{0,1\}^N} p(x)=1 \Big\} . 
\end{equation*}

\paragraph{Hidden units, visible marginal distributions}

We will be interested in a situation where only a subset $V\subseteq N$ of all units can be observed, while the other units $H=N\setminus V$ are unobserved or hidden. 
Given the probability distribution $p(x;\theta)$ over the states $x = (x_V,x_H)\in\{0,1\}^V\times \{0,1\}^H$ of all units, 
the marginal probability distribution over the visible states $x_V$ is given by 
\begin{equation}
p(x_V;   \theta) = \sum_{x_H\in \{0,1\}^H} p(x;\theta),\quad  x_V\in\{0,1\}^V. 
\label{eq:defrbm} 
\end{equation}
The set of marginal probability distributions, for all choices of $\theta$, is a subset of the $2^V-1$ dimensional simplex $\Delta_{\{0,1\}^V}$. 
It is the image of the fully observable Boltzmann machine probability manifold by the linear map that computes marginal distributions. 
In general this set is no longer a manifold. It may have a rather complex shape with self intersections and dimension strictly smaller than that of the manifold of distributions of all units. 
We will be concerned with the properties of this set in the special case where interaction edges are only allowed between visible and hidden units.

\section{Restricted Boltzmann machines}
\label{sec:definitions}

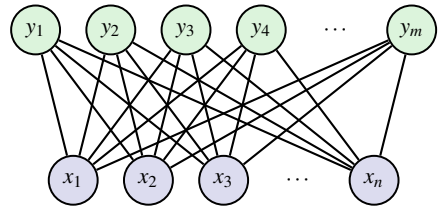
\begin{figure}
	\centering
	\tikzset{node distance=2cm, auto}
	\begin{tikzpicture}[scale=1, every node/.style={transform shape}]
	
	\tikzstyle{neuronbl}=[circle, line width=.75pt, draw=black, inner sep=.025cm, minimum size = .65cm, fill=bl!15]	
	\tikzstyle{neurongr}=[circle, line width=.75pt, draw=black, inner sep=.025cm, minimum size = .65cm, fill=gr!15]	
	
	\node[neuronbl] (I-1) at (1,0) {$x_{1}$}; 
	\node[neuronbl] (I-2) at (2,0) {$x_{2}$}; 
	\node[neuronbl] (I-3) at (3,0) {$x_{3}$}; 
	\node[] 		(I-d) at (4,0) {$\cdots$};
	\node[neuronbl] (I-4) at (5,0) {$x_{n}$}; 
	
	
	\node[neurongr] (H-1) at (1 -.5,2) {$y_{1}$}; 
	\node[neurongr] (H-2) at (2 -.5,2) {$y_{2}$}; 
	\node[neurongr] (H-3) at (3 -.5,2) {$y_{3}$}; 
	\node[neurongr] (H-4) at (4 -.5,2) {$y_{4}$}; 
	\node[] 		(H-d) at (5 -.5,2) {$\cdots$};
	\node[neurongr] (H-5) at (6 -.5,2) {$y_{m}$};
	
	
	\draw[-, line width = .8pt] (H-1) edge node[left] {} (I-1);
	\draw[-, line width = .8pt] (H-1) edge (I-2);
	\draw[-, line width = .8pt] (H-1) edge (I-3);
	\draw[-, line width = .8pt] (H-1) edge (I-4);
	
	\draw[-, line width = .8pt] (H-2) edge (I-1);
	\draw[-, line width = .8pt] (H-2) edge (I-2);
	\draw[-, line width = .8pt] (H-2) edge (I-3);
	\draw[-, line width = .8pt] (H-2) edge (I-4);
	
	\draw[-, line width = .8pt] (H-3) edge (I-1);
	\draw[-, line width = .8pt] (H-3) edge (I-2);
	\draw[-, line width = .8pt] (H-3) edge (I-3);
	\draw[-, line width = .8pt] (H-3) edge (I-4);
	
	\draw[-, line width = .8pt] (H-4) edge (I-1);
	\draw[-, line width = .8pt] (H-4) edge (I-2);
	\draw[-, line width = .8pt] (H-4) edge (I-3);
	\draw[-, line width = .8pt] (H-4) edge (I-4);
	
	\draw[-, line width = .8pt] (H-5) edge (I-1);
	\draw[-, line width = .8pt] (H-5) edge (I-2);
	\draw[-, line width = .8pt] (H-5) edge (I-3);
	\draw[-, line width = .8pt] (H-5) edge node[right] {} (I-4);
	\end{tikzpicture}
	\caption{RBM as a graphical model with visible units $x_1,\ldots, x_n$ and hidden units $y_1,\ldots,y_m$. 
	Each edge has an associated interaction weight $w_{ji}$, each visible node has an associated bias weight $b_i$, and each hidden node an associated bias weight $c_j$. }
	\label{fig:RBM}
\end{figure}

The restricted Boltzmann machine (RBM) is a special type of Boltzmann machine where the interactions are restricted to be between visible and hidden units, such that $I=\{\{ i, j\} \colon i\in V, j\in H \}$.   
This is illustrated in Figure~\ref{fig:RBM}. 
The corresponding probability distributions take the form 
\begin{equation}
p(x;\theta) = \frac{1}{Z(\theta)}\sum_{y\in\{0,1\}^H}\exp(y^\top W x + c^\top y + b^\top x), \quad x\in\{0,1\}^V. 
\label{eq:rbm}
\end{equation}
Here $x$ is the state of the visible units, $y$ is the state of the hidden units, $Z$ is the partition function, 
and $\theta = (W,b,c)$ denotes the parameters, composed of the interaction weights $W=(w_{j,i})_{j\in H, i\in V}$, the biases of the visible units $b=(b_i)_{i\in V}$, and the biases of the hidden units $c=(c_j)_{j\in H}$. 
The RBM probability model with $n$ visible and $m$ hidden units is the set of probability distributions of the form~\eqref{eq:rbm}, for all possible choices of $\theta$. We denote this set by $\RBM_{n,m}$. 
We will write $[n]=\{1,\ldots, n\}$ and $[m]=\{1,\ldots, m\}$ to enumerate the visible and hidden units, respectively. 
We write $\Xcal=\{0,1\}^V$ for the state space of the visible units, and $\Ycal=\{0,1\}^H$ for that of the hidden units. 

An RBM probability model can be interpreted in various interesting and useful ways, as we discuss in the following. 
These are views of the same object and are equivalent in that sense, but they highlight different aspects.

\paragraph{Product of mixtures}

One interpretation the RBM is as a \emph{product of experts} model, meaning that it consists of probability distributions which are normalized entrywise products with factors coming from some fixed models. 
Factorized descriptions are familiar from graphical models, where one considers probability distributions that factorize into potential functions, which are arbitrary positive valued functions that depend only on certain fixed subsets of all variables. 
We discuss graphical models in more depth in Section~\ref{sec:hierarchicalmodels}. 
In the case of RBMs, each factor model is given by mixtures of product distributions. 
A \emph{product distribution} is a distribution of multiple variables which factorizes as an outer product $q(x_1,\ldots, x_n)=\prod_{i\in [n]}q_i(x_i)$ of distributions $q_i$ of the individual variables. 
A \emph{mixture distribution} is a convex combination $q(x)=\sum_k\lambda_k q_k(x)$, where the $\lambda_k$ are non-negative weights adding to one, and the $q_k$ are probability distributions from some given set. 
Indeed, the RBM distributions can be written as 
\begin{align}
p(x;\theta) 
=& \frac{1}{Z(\theta)} \sum_{y\in\{0,1\}^m}\exp(y^\top W x + c^\top y + b^\top x) \nonumber \\
=& \frac{1}{Z(\theta)} \exp(b^\top x) \prod_{j\in [m]} (1 + \exp(W_{j:} x + c_j)) \\
=& \frac{1}{Z(\theta)} \prod_{j\in [m]} (\exp(W_{j:}' x) + \exp(c_j) \exp(W_{j:}'' x)) . \nonumber
\end{align}
Here  $W_{j:}'$ and $W_{j:}'' = W_{j:}+W_{j:}'$ can be chosen arbitrarily in $\mathbb{R}^n$ for all $j\in [m]$, with $b = \sum_{j\in [m]} W_{j:}'$. 
In turn, for any mixture weights $\lambda_j\in(0,1)$ we can find suitable $c_j\in\mathbb{R}$, and for any distributions $p_{j,i}'$ and $p_{j,i}''$ on $\Xcal_i=\{0,1\}$ suitable $W_{j,i}'$ and $W_{j,i}''$, such that 
\begin{align}
p(x;\theta) 
=& \frac{1}{Z(\theta)} \prod_{j\in [m]} \Big(\lambda_j \prod_{i\in [n]}p_{j,i}'(x_i) + (1-\lambda_j) \prod_{i\in [n]}p_{j,i}''(x_i)\Big). 
\label{eq:prodmix}
\end{align}
This shows that the RBM model can be regarded as the set distributions that are entrywise products of $m$ terms, with each term being a mixture of two product distributions over the visible states. 

Products of experts can be trained in an efficient way, with methods such as contrastive divergence, which we will outline in Section~\ref{sec:training}. 
Products of experts also relate to the notion of distributed representations, where each observation is explained by multiple latent causes. 
This allows RBMs to create exponentially many inference regions, or possible categorizations of input examples, on the basis of only a polynomial number of parameters. 
This sets RBMs apart from mixture models, and provides one way of breaking the curse of dimensionality, which is one motivation for choosing one network architecture over another in the first place. We discuss more about this further below and in Section~\ref{sec:representational}.

\paragraph{Tensors and polynomial parametrization}

A probability distribution on $\{0,1\}^n$ can be regarded as an $n$-way table or tensor with entries indexed by $x_i \in\{0,1\}$, $i\in [n]$. 
A tensor $p$ is said to have rank one if it can be factorized as $p=p_1\otimes \cdots\otimes p_n$, where each $p_i$ is a vector. 
Thus, non-negative rank one tensors correspond to product distributions. 
A tensor is said to have non-negative rank $k$ if it can be written as the sum of $k$ non-negative tensors of rank $1$, and $k$ is the smallest number for which this is possible. 
Tensors of non-negative rank at most $k$ correspond to mixtures of~$k$ product distributions. 
The RBM distributions are, up to normalization, the tensors that can be written as Hadamard (i.e., entrywise) products of $m$ factor tensors of non-negative rank at most two. 
The representable tensors have the form 
\begin{equation}
p = \prod_{j\in [m]} \big(q'_{j,1}\otimes \cdots\otimes q'_{j,n} + q''_{j,1}\otimes \cdots\otimes q''_{j,n}\big), 
\end{equation}
where the $q'_{j,i}$ and $q''_{j,i}$ are non-negative vectors of length two. 

In particular, we note that, up to normalization, the RBM distributions have a polynomial parametrization 
\begin{equation}
p =  \Big(\prod_{i\in [n]}\omega_{0,i}^{x_i}\Big) \prod_{j\in [m]} \Big(1 + \omega_{j,0} \prod_{i\in [n]}\omega_{j,i}^{x_i} \Big), 
\end{equation}
with parameters $\omega_{0,i}=\exp(b_i)\in \mathbb{R}_+$, $\omega_{j,0}=\exp(c_j)\in\mathbb{R}_+$, $j\in [m]$,  $\omega_{j,i}=\exp(W_{j,i})\in\mathbb{R}_+$, $(i,j)\in [n]\times [m]$. 
The fact that RBMs have a polynomial parametrization makes them, like many other probability models, amenable to be studied with tools from algebra. 
This is the realm of \emph{algebraic statistics}. Introductions to this area at the intersection of mathematics and statistics are~\cite{drton2009lectures,sullivant2018}. 
In algebraic geometry one studies questions such as the dimension and degree of solution sets of polynomial equations. 
When translated to statistics, these questions relate to parameter identifiability, the number of maximizers of the likelihood function, and other important properties of statistical models. 

\paragraph{Kronecker products, harmonium models}

As we have seen, the joint distributions of a Boltzmann machine form an exponential family over the states of all units. 
That is, the joint distributions are given by exponentiating and normalizing vectors from an affine space,  
\begin{equation}
p(x,y ;\theta) = \frac{1}{Z(\theta)}\exp(\theta^\top F(x,y)), \quad (x,y)\in\Xcal\times\Ycal. 
\end{equation}
Here the sufficient statistics $F_1,\ldots, F_d\colon \Xcal\times\Ycal\to \mathbb{R}$ span the affine space in question. 
For an RBM, the sufficient statistics $F$ have a special structure. 
Recall that the Kronecker product of two matrices is defined by $(a_{i,j})_{i,j}\otimes (b_{k,l})_{k,l}=(a_{i,j} (b_{k,l})_{k,l})_{i,j}=(a_{i,j} b_{k,l})_{(i,k),(j,l)}$. 
The sufficient statistics for the exponential family of the RBM can be written as a Kronecker product 
\begin{equation}
F(x,y)=F^V(x) \otimes F^H(y),  \quad (x,y)\in\Xcal\times\Ycal, 
\end{equation} 
where $F^V(x)=(1, x_1,\ldots, x_n)^\top$ and $F^H(y)=(1,y_1,\ldots,y_m)^\top$ are sufficient statistics of the independence models of the $n$ visible binary units and the $m$ hidden binary units. 
The independence model is the exponential family of product distributions, $\frac{1}{Z}\exp(\sum_i \theta_i F_i^V(x)) = \frac{1}{Z}\exp(w^\top x + c)=\frac{1}{Z}\prod_{i\in[n]}\exp(w_i x_i)$. 
%

\begin{figure}
	\centering
\includegraphics[scale=.7]{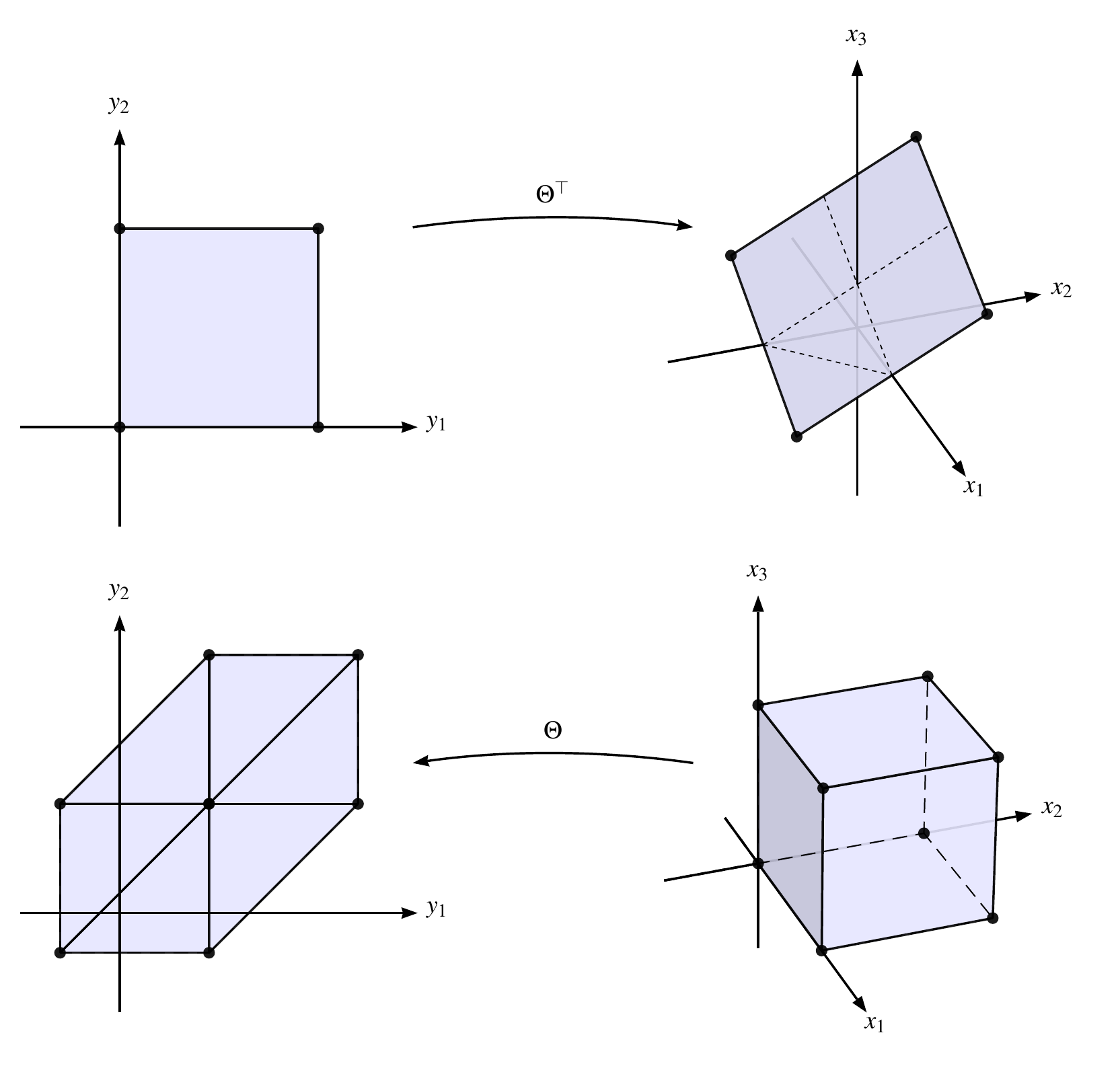}
\caption{
For an RBM, the conditional distributions $p(X|y;\theta)$ of the visible variables given the hidden variables, are the elements of an exponential family with sufficient statistics $F^V$ and parameters given by projections $\Theta^\top F^H(y)$ of the sufficient statistics $F^H$ of the hidden variables. 
Similarly, $p(Y|x;\theta)$ are exponential family distributions with sufficient statistics $F^H$ and parameters $\Theta F^V(x)$. 
The figure illustrates these vectors for $\RBM_{3,2}$ and a choice of $\theta$. }
\label{fig:condRBM32}
\end{figure}

The Kronecker product structure allows us to express the conditional distribution of hidden units given visible units, and vice versa, in the following simple way. 
Given two vectors $a,b$, write $\langle a, b\rangle$ for their inner product $a^\top b = \sum_i a_i b_i$. 
Take any parameter vector $\theta\in\mathbb{R}^{(n+1)(m+1)}$ and arrange its entries into a matrix $\Theta\in\mathbb{R}^{(m+1)\times (n+1)}$, going column by column.  
Then 
\begin{eqnarray*}
\big\langle\theta, F(x,y) \big\rangle 
&=& \big\langle\theta, F^V(x) \otimes  F^H(y)\big\rangle \\
&=& \big\langle \Theta^\top  F^H(y), F^V(x) \big\rangle \\
&=& \big\langle \Theta F^V(x), F^H(y)\big\rangle . 
\label{eq:Roth}
\end{eqnarray*}  
%
These expression describe following probability distributions: 
\begin{eqnarray*}
p(x,y;\theta) &=& \frac{1}{Z(\theta)} \exp\big(\big\langle\theta, F(x,y) \big\rangle\big) \\
p(x|y;\theta) &=& \frac{1}{Z\big(\Theta^\top  F^H(y)\big)} \exp\big(\big\langle \Theta^\top  F^H(y), F^V(x)  \big\rangle\big) \\
p(y|x;\theta) &=& \frac{1}{Z\big(\Theta F^V(x) \big)} \exp\big(\big\langle \Theta F^V(x), F^H(y) \big\rangle\big).
\end{eqnarray*}
Geometrically, $\Theta F^V$ is a linear projection of $F^V$ into the parameter space of the exponential family with sufficient statistics $F^H$ and, similarly, $\Theta^\top  F^H$ is a linear projection of $F^H$ into the parameter space of an exponential family for the visible variables. 
This is illustrated in Figure~\ref{fig:condRBM32}.

\paragraph{Restricted mixtures of products}

The marginal distributions can always be written as 
$$
p(x;\theta) = \sum_{y} p(x,y;\theta) = \sum_{y} p(y;\theta) p(x|y;\theta),\quad x\in\Xcal. 
$$
In the case of an RBM, the conditional distributions are product distributions $p(x|y;\theta)=\prod_{i\in [n]}p(x_i|y;\theta)$. 
In turn, the RBM model consists of mixtures of product distributions, with mixture weights $p(y;\theta)$. 
However, the marginal $p(y;\theta)$ and the tuple of conditionals $p(x|y;\theta)$ have a specific and constrained structure. 
For instance, as can be seen in Figure~\ref{fig:condRBM32} for the model $\RBM_{3,2}$, the mixture components have parameter vectors that are affinely dependent. 
One implication is that $\RBM_{3,2}$ cannot represent any distribution with large values on the even parity strings $000, 011, 101, 110$ and small values on the odd parity strings $001, 010, 100, 111$. 
This kind of constraint, coming from constraints on the mixture components, have been studied in~\cite{montufar2015does}. 
An exact description of the constraints that apply to the probability distributions within $\RBM_{3,2}$ was obtained recently in~\cite{SeigalMontufar}. 
We comment on this later in Section~\ref{sec:semialgebraic}.

\paragraph{Superposition of soft-plus units} 

Another useful way of viewing RBMs is as follows. 
The description as products of mixtures shows that in RBMs the log-probabilities are sums of independent terms. 
More precisely, they are superpositions of $m$ soft-plus units and one linear unit: 
\begin{equation}
\log( p(x;\theta) ) = 
\sum_{j\in[m]} \log(1 + \exp(W_{j:} x + c_j)) 
+  b^\top x  - \log(Z(\theta)). 
\label{eq:softplus}
\end{equation}
A \emph{soft-plus unit} computes a real valued affine function of its arguments, $x\mapsto w^\top x +c$, and then applies the soft-plus non linearity $s\mapsto \log(1 + \exp(s))$. 
A linear unit simply computes $x\mapsto b^\top x + c$. 

Log-probabilities correspond uniquely to probability distributions. 
When studying the space of representable log-probabilities, it is helpful to allow ourselves to add or disregard additive constants, since they correspond to scaling factors that cancel out with the normalization of the probability distributions. 

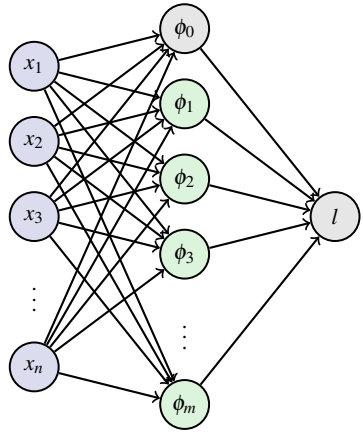
\begin{figure}
	\centering
	\tikzset{node distance=2cm, auto}
	\begin{tikzpicture}[scale=1, every node/.style={transform shape}]
	
	\tikzstyle{neuronbl}=[circle, line width=.75pt, draw=black, inner sep=.025cm, minimum size = .65cm, fill=bl!15]	
	\tikzstyle{neurongr}=[circle, line width=.75pt, draw=black, inner sep=.025cm, minimum size = .65cm, fill=gr!15]	
	\tikzstyle{neuronout}=[circle, line width=.75pt, draw=black, inner sep=.025cm, minimum size = .65cm, fill=black!10]	
	\tikzstyle{neuronout2}=[circle, line width=.75pt, draw=black, inner sep=.025cm, minimum size = .65cm, fill=white]	
	
	\node[neuronbl] (I-1) at (0,4) {$x_{1}$}; 
	\node[neuronbl] (I-2) at (0,3) {$x_{2}$}; 
	\node[neuronbl] (I-3) at (0,2) {$x_{3}$}; 
	\node[]         (I-d) at (0,1) {$\vdots$};
	\node[neuronbl] (I-4) at (0,0) {$x_{n}$}; 
	
	\node[neuronout] (O-1) at (2,4.5) {$\phi_0$}; 
	\node[neurongr] (O-2) at (2,3.5) {$\phi_1$}; 
	\node[neurongr] (O-3) at (2,2.5) {$\phi_2$}; 
	\node[neurongr] (O-4) at (2,1.5) {$\phi_3$}; 
	\node[]          (O-d) at (2,.5) {$\vdots$};
	\node[neurongr] (O-5) at (2,-.5) {$\phi_{m}$}; 
	
	\node[neuronout] (F) at (4,2) {$l$}; 
	
	\draw[->, line width = .8pt] (I-1) edge node[above] {} (O-1);
	\draw[->, line width = .8pt] (I-2) edge (O-1);
	\draw[->, line width = .8pt] (I-3) edge (O-1);
	\draw[->, line width = .8pt] (I-4) edge (O-1);
	
	\draw[->, line width = .8pt] (I-1) edge (O-2);
	\draw[->, line width = .8pt] (I-2) edge (O-2);
	\draw[->, line width = .8pt] (I-3) edge (O-2);
	\draw[->, line width = .8pt] (I-4) edge (O-2);
	
	\draw[->, line width = .8pt] (I-1) edge (O-3);
	\draw[->, line width = .8pt] (I-2) edge (O-3);
	\draw[->, line width = .8pt] (I-3) edge (O-3);
	\draw[->, line width = .8pt] (I-4) edge (O-3);
	
	\draw[->, line width = .8pt] (I-1) edge (O-4);
	\draw[->, line width = .8pt] (I-2) edge (O-4);
	\draw[->, line width = .8pt] (I-3) edge (O-4);
	\draw[->, line width = .8pt] (I-4) edge (O-4);
	
	\draw[->, line width = .8pt] (I-1) edge (O-5);
	\draw[->, line width = .8pt] (I-2) edge (O-5);
	\draw[->, line width = .8pt] (I-3) edge (O-5);
	\draw[->, line width = .8pt] (I-4) edge node[below] {} (O-5);

	\draw[->, line width = .8pt] (O-1) edge (F);
	\draw[->, line width = .8pt] (O-2) edge (F);
	\draw[->, line width = .8pt] (O-3) edge (F);
	\draw[->, line width = .8pt] (O-4) edge (F);
	\draw[->, line width = .8pt] (O-5) edge (F);
	
	\end{tikzpicture}
	\caption{An RBM model can be regarded as the set of log-probabilities which are computable as the sum of a linear unit $\phi_0$ and $m$ soft-plus units $\phi_j$, $j=1,\ldots, m$. }
	\label{fig:soft-plus}
\end{figure}

The RBM model can be regarded as the set of negative energy functions (log-probabilities modulo additive constants) that can be computed by a feedforward network with one hidden layer of $m$ soft-plus units and one linear unit, and a single output unit adding the outputs of the hidden units. 
The situation is illustrated in Figure~\ref{fig:soft-plus}.  
Feedforward networks are often conceptually easier than stochastic networks or probability graphical models. 
One point to note is that the output unit of the RBM energy network  only computes unweighted sums. 

A type of computational unit that is closely related to the soft-plus unit is the \emph{rectified linear unit} (ReLU). A ReLU computes a real valued affine function of its arguments, $x\mapsto w^\top x +c$, followed by rectification $s\mapsto [s]_+ = \max\{0,s\}$. 
As it turns out, if we replace the soft-plus units by ReLUs in eq.~\eqref{eq:softplus}, we obtain the so-called tropical RBM model, which is a piecewise linear version of the original model that facilitates a number of computations. 
We discuss more details of this relationship in the next paragraph.

\paragraph{Tropical RBM, superposition of ReLUs}

The tropical RBM model is the set of vectors that we obtain when evaluating log-probabilities of the RBM model using the max-plus algebra and disregarding additive constants. 
We replace sums by maximum, so that a log-probability vector $l(x;\theta)  = \sum_{y}\exp(y^\top W x + b^\top x + c^\top y)$, $x\in \Xcal$, becomes $\Phi(x;\theta) = \max_y \{  y^\top W x + b^\top x + c^\top y \}$, $x\in\Xcal$. 
We can write this more compactly as 
\begin{equation}
\Phi(x;\theta) = \theta^\top F(x, h(x;\theta)), \quad x\in\Xcal, 
\end{equation}
where $F(x,y)=(1,x_1,\ldots, x_n)^\top\otimes(1,y_1,\ldots, y_m)^\top$ is the vector of sufficient statistics, and $h(x;\theta) = \argmax_y \theta^\top F(x, y)= \argmax_y p(y|x;\theta)$ is the \emph{inference function} that returns the most probable $y$ given $x$. 
In particular, the tropical RBM model is the image of a piecewise linear map. 

We note the following decomposition, which expresses the tropical RBM model as a superposition of one linear unit and $m$ ReLUs. 
We have 
\begin{eqnarray*}
	\Phi(x;\theta) 
	&=& \max_{y} \{  y^\top W x + b^\top x + c^\top y \}\\
	&=&  b^\top x + \sum_{j\in[m]} \max_{y_j}\{y_j W_{j:} x + c_j y_j\}\\
	&=&  b^\top x + \sum_{j\in [m]} [ W_{j:}x + c_j]_+.  
\end{eqnarray*} 
In turn, the tropical RBM is the set of vectors computable by a sum of one linear unit $x\mapsto b^\top x$ and $m$ ReLUs $x\mapsto [w^\top x + c]_+  = \max\{ 0, w^\top x +c \}$. 

\begin{figure}
	\centering
	\includegraphics[clip=true, trim=1cm 1.5cm 1cm 2cm,scale=.7]{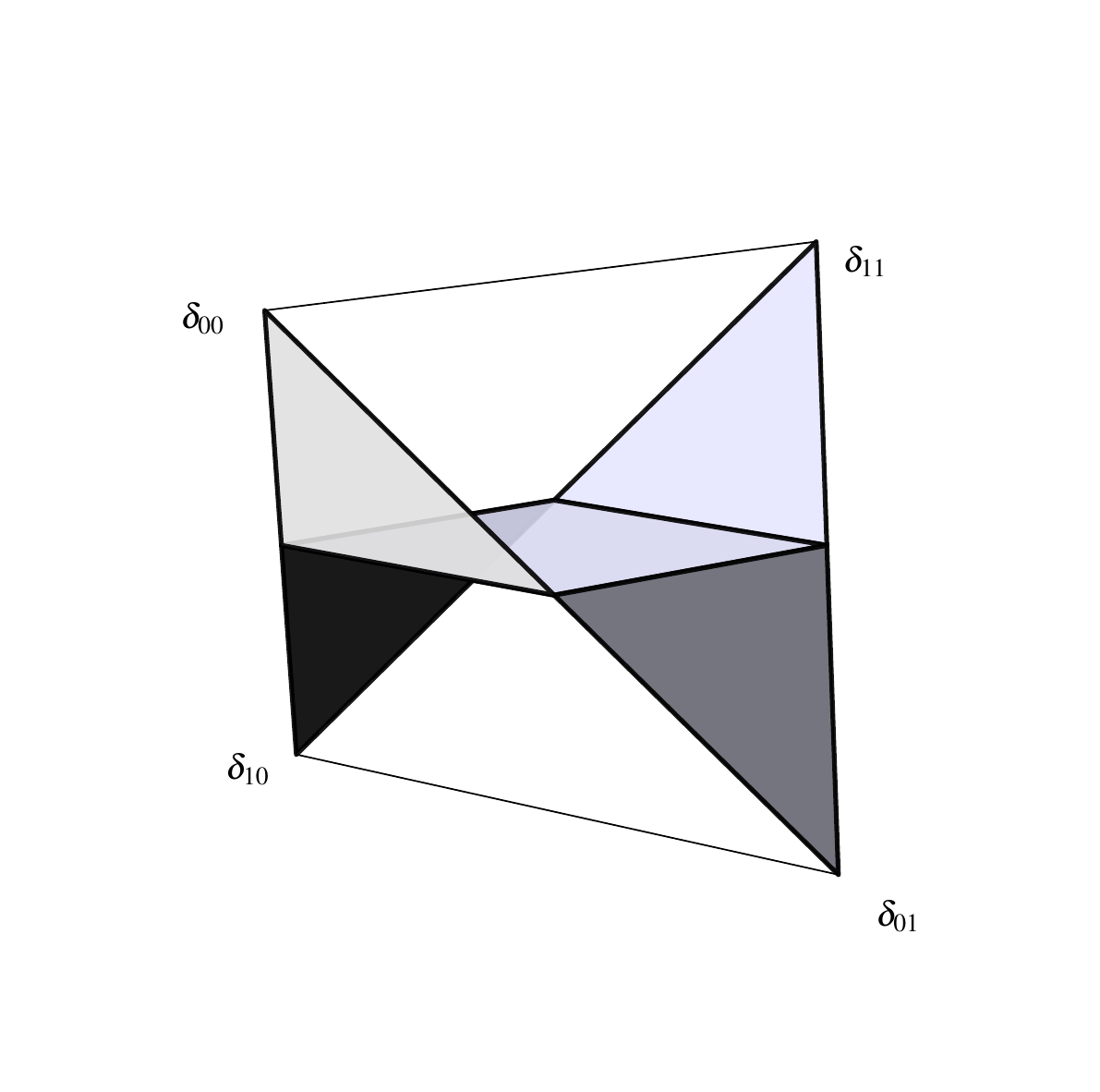}
	\caption{Illustration of the set of functions 
		$([w^\top x + c]_+)_x$, $x\in\{0,1\}^2$, that can be represented by a ReLU with two binary inputs. This corresponds to the tropical RBM model with zero biases on the visible units. 
		For the visualization of this $3$ dimensional set in $\mathbb{R}^4_{\geq0}$, we scaled the vectors to have entry sum $1$ (the zero function is identified with the one function), which results in the shown subset of the simplex with vertices $\delta_{x}$  the indicators of individual inputs $x\in\{0,1\}^2$. 
	}
	\label{fig:ReLU}
\end{figure}

The set of functions that can be represented by a ReLU is closed under multiplication by non-negative scalars. 
Hence the unweighted sums of $m$ ReLUs, 
$\sum_{j\in [m]} [w_j^\top x + c_j]_+$, 
express the same set of functions as the conic combinations of $m$ ReLUs, 
$\sum_{j\in[m]} \alpha_j [\bar w_j^\top x + \bar c_j]_+$, where $\alpha_j\geq 0$, $j\in[m]$. 
For analysis and visualization, we can disregard positive multiplicative factors, 
and consider convex combinations of $m$ normalized ReLUs. 
We can normalize each function such that its entry sum equals one. 
Zero functions cannot be normalized in this way, but they are equivalent to constant functions. 
The set of normalized functions expressible by a ReLU with two binary inputs is shown in Figure~\ref{fig:ReLU}. 
A sum of $m$ ReLUs can realize any convex combinations of $m$ points from this set. 
Affine functions with positive values correspond to the horizontal square in the middle of the figure,  and constant functions to the point at the center of the square. 
Adding positive / negative constants to a given point corresponds to moving from it towards / away from the center. 
%

\paragraph{Other generalizations}

There are numerous generalizations of the regular RBM model. 
\begin{itemize}
\item 
A Boltzmann machine can be defined with discrete non-binary states, real valued Gaussian units, or any other type of probability model for each unit. 
If the hidden variables are defined to take $k$ possible values each, then the RBM defines a Hadamard product of tensors of non-negative rank at most $k$~\cite{montufar2013discrete}. 
In particular, this is a generalization of mixtures of products models. 
Visible units with more than two states have been used, for example, in collaborative filtering~\cite{Salakhutdinov:2007}. 

\item 
Viewed as Kronecker product models, 
with distributions $\frac{1}{Z(\theta)}\sum_y\exp( \theta^\top F^V(x)\otimes F^H(y))$, 
RBMs can be generalized to have arbitrary factors $F^V$ and $F^H$, rather than just sufficient statistics of independence models. 
In this case, the conditional distributions of the visible variables, given the hidden variables, are distributions from the exponential family specified by $F^V$. 
This setting has been discussed in~\cite{montufar2015dimension} and in~\cite{welling:exponential} by the name \emph{exponential family harmonium}. 

\item 
We can extend the setting of pair interactions to models with higher order interactions, called higher order Boltzmann machines~\cite{Sejnowski86higher-orderboltzmann}. 

\item 
Other generalizations include deep architectures, such as deep belief networks~\cite{Hinton:2006:FLA:1161603.1161605} and deep Boltzmann machines~\cite{SalHinton07}. 
Here one considers a stack of layers of units, with interactions restricted to pairs of units at adjacent layers. 
The representational power of deep belief networks has been studied in~\cite{DBLP:journals/neco/SutskeverH08,LeRoux:2008:RPR:1374176.1374187,montufar2011refinements,montufar2014universal} and that of deep Boltzmann machines in~\cite{montufar2014deep}. 
\item 
For some applications, such as discriminative tasks, structured output prediction, stochastic control, one splits the visible units into a set of inputs and a set of outputs. 
The representational power of \emph{conditional RBMs} has been studied in~\cite{montufar2014expressive}. 
\item 
Another line of generalizations are quantum models~\cite{PhysRevX.8.021050}. 
\item A recent overview on RBM variants for diverse applications was given in~\cite{ZHANG20181186}. 
\end{itemize}

\section{Basics of training}
\label{sec:training}

We give a short introduction to training. 
The general idea of training is to adjust the parameters of the Boltzmann machine such that it behaves in a desirable way. 
To do this, we first decide on a function to measure the desirability of the different possible behaviors, and then maximize that function over the model parameters. 
The first explicit motivation and derivation of a learning algorithm for Boltzmann machines is by Ackley, Hinton, and Sejnowski~\cite{Ackley85alearning}, based on statistical mechanics. Given a set of examples, the algorithm modifies the interaction weights and biases of the network so as to construct a generative model that produces examples with the same probability distribution of the provided examples. 

\paragraph{Maximizing the likelihood of a data set} 

Based on a set of examples, we aim at generating examples with the same probability distribution. 
To this end, we can maximize the log-likelihood of the provided examples with respect to the Boltzmann machine model parameters. 
For a set of examples $x^1,\ldots, x^N \in \{0,1\}^n$, the log-likelihood is 
\begin{equation}
L(\theta) = \sum_{i=1}^N \log p(x^i;\theta) = \sum_{x} p_{\text{data}}(x) \log p(x;\theta), 
\label{eq:likelihood}
\end{equation}
where $p_{\text{data}}$ is the empirical data distribution $p_{\text{data}}(x) = \frac{1}{N}\sum_{i=1}^N \delta_{x^i}(x)$, $x\in\Xcal$, and $p(x;\theta)$, $x\in\Xcal$, is the model distribution with parameter $\theta\in\mathbb{R}^d$. 
Maximizing~\eqref{eq:likelihood} with respect to $\theta$ is equivalent to minimizing the Kullback-Leibler divergence $D(p_{\text{data}} \| p_\theta)$ from $p_{\text{data}}$ to the model distribution $p_\theta\equiv p(\cdot;\theta)$, again with respect to $\theta$. 
The divergence is defined as 
\begin{equation}
D(p_{\text{data}} \| p_\theta) = \sum_{x} p_{\text{data}}(x) \log\frac{p_{\text{data}}(x)}{p(x;\theta)}. 
\end{equation} 
In some cases the minimum might not be attained by any value of the parameter $\theta$. 
However, it is attained as $D(p_{\text{data}}\| p)$ for some distribution $p$ in the closure of $\{p_\theta\colon \theta\in\mathbb{R}^d\}\subseteq\Delta_{\Xcal}$.

\paragraph{Likelihood gradient}

In most cases, we do not know how to maximize the log-likelihood in closed form (we discuss a recent exception to this in Section~\ref{sec:semialgebraic}). 
We can search for a maximizer by initializing the parameters at some random value $\theta^{(0)}$ and iteratively adjusting them in the direction of the gradient, as 
\begin{equation}
\theta^{(t+1)}=\theta^{(t)} + \alpha_t \nabla L(\theta^{(t)}),  
\end{equation}
until some convergence criterion is met. 
Here the \emph{learning rate} $\alpha_t>0$ is a hyper-parameter of the learning criterion that needs to be specified. 
Typically the user tries a range of values. 
Often in practice, the parameter updates are computed based only on subsets of the data at the time, in what is known as on-line, mini-batch, or stochastic gradient. 

Writing $F\colon \Xcal\times\Ycal\to \mathbb{R}^d$ for the sufficient statistics of an exponential family of joint distributions of visible and hidden variables, we have 
\begin{equation}
\nabla L(\theta) 
= \langle F \rangle_{\text{data}}   - \langle F \rangle_{\theta} . 
\label{eq:likelihoodgrad}
\end{equation}
Here $\nabla = (\frac{\partial}{\partial\theta_1}, \ldots,\frac{\partial}{\partial\theta_d})^\top$ is the column vector of partial derivatives with respect to the model parameters,  
$\langle\cdot\rangle_{\text{data}}$ stands for the expectation value with respect to the joint probability distribution $p_{\text{data}}(x)p_\theta(y|x)$, and 
$\langle\cdot\rangle_{\theta}$ stands for the expectation with respect to the joint distribution $p_\theta(x,y)$.

The computation of the gradient can be implemented as follows. 
We focus on the binary RBM, for which the sufficient statistics take the form  
$$
F(x,y)=(F_I, F_V, F_H)(x,y) = (( y_jx_i )_{j\in H,i\in V} , (x_i)_{i\in V}, (y_j)_{j\in H}), \; (x,y)\in\{0,1\}^V\times\{0,1\}^H. 
$$ 

For the expectation value in~\eqref{eq:likelihoodgrad} involving the data distribution: 
\begin{itemize}
\item 
Write a data matrix $\tilde X  = ( x^1, \cdots, x^N)$. 
\item 
Collect the activation probabilities of the individual hidden units, in response to each visible data vector, into a matrix $\tilde Y = \sigma( c \cdot  \mathds{1}_{1\times N} + W \cdot \tilde X)$. 
Here $\sigma$ is the logistic function $s\mapsto 1/(1 + \exp(-s))$ applied entrywise to the argument, and $\mathds{1}_{1\times N}$ is the $1\times N$ matrix of ones. 
\item 
Then 
\begin{eqnarray}
\langle F_I\rangle_{\text{data}} &=& \tilde Y \cdot  \tilde X^\top / N,  \nonumber \\
\langle F_V\rangle_{\text{data}} &=& \tilde X \cdot  \mathds{1}_{N\times 1}/N, \label{eq:datasam}\\
\langle F_H\rangle_{\text{data}} &=& \tilde Y \cdot  \mathds{1}_{N\times 1}/N \nonumber. 
\end{eqnarray}
\end{itemize}
This calculation is relatively tractable, with order $Nnm$ operations. 

For the expectation in~\eqref{eq:likelihoodgrad} with respect to the model distribution: 
\begin{itemize}
\item 
Write $X$ for the matrix with columns all vectors in $\{0,1\}^n$ 
and $Y$ for the matrix with columns all vectors in $\{0,1\}^m$. 
\item 
Let $P_{Y\times X}$ be the matrix with entries $p_\theta(x,y)$, with rows and columns indexed by $y$ and $x$. 
\item 
Then 
\begin{eqnarray}
\langle F_I\rangle_{\theta} &=& Y \cdot  P_{Y\times X} \cdot  X^\top, \nonumber \\
\langle F_V\rangle_{\theta} &=&  \mathds{1}_{1\times 2^m}\cdot P_{Y\times X} \cdot X^\top, \\
\langle F_H\rangle_{\theta} &=&  Y \cdot  P_{Y\times X} \cdot\mathds{1}_{2^n\times 1} . \nonumber 
\end{eqnarray}
\end{itemize}
This calculation is possible for small models, 
but it can quickly become intractable. Since $P_{Y\times X}$ has $2^m$ rows and $2^n$ columns, computing its partition function and the expectations requires exponentially many operations in the number of units. 
In applications $n$ and $m$ may be in the order of hundreds or thousands. 
In order to overcome the intractability of this computation, a natural approach is to approximate the expectation values by sample averages. 
We discuss this next.

\paragraph{Contrastive divergence}

The expectations $\langle F\rangle_{\theta}$ with respect to the model distribution can be approximated in terms of sample averages obtained by Gibbs sampling the RBM. 
One method based on this idea is \emph{contrastive divergence} (CD)~\cite{Hinton2002}. This method has been enormously valuable in practical applications and is the standard learning algorithm for RBMs. 
The CD algorithm can be implemented as follows. 
\begin{itemize}
\item As before, write a data matrix $\tilde X = (x^1,\ldots, x^N)$. 
\item Then update the state of the hidden units of the RBM by 
$$
\tilde Y = ( \sigma(c\cdot \mathds{1}_{1\times N} + W\cdot \tilde X)\geq \operatorname{rand}_{m\times N}).
$$  
\item 
Update the state of the visible units by 
$$
\hat X = (\sigma(b\cdot \mathds{1}_{1\times N} + W^\top \tilde Y ) \geq \operatorname{rand}_{n\times N}).
$$ 
These updates are the Gibbs sampling state updates described in eq.~\eqref{eq:transitionprob}, computed in parallel for all hidden and visible units. 
Here $\operatorname{rand}_{n\times N}$ is an $n\times N$ array of independent variables uniformly distributed in $[0,1]$, and $\geq$ is evaluated entrywise as a logic gate with binary outputs.  
\item 
Now use the reconstructed data $\hat X$ 
to compute $\langle F\rangle_{\text{recon}}$ in the same way as $\tilde X$ was used to compute $\langle F\rangle_{\text{data}}$ in eq.~\eqref{eq:datasam}. 
The approximate model sample average $\langle F\rangle_{\text{recon}}$ 
is then used as an approximation of $\langle F\rangle_{\theta}$. 
\end{itemize}
This calculation involves only order $Nnm$ operations, and remains tractable even for relatively large $n$ and $m$ in the order of thousands. 

CD is an approximation to the maximum likelihood gradient. 
The bias of this method with respect to the actual gradient has been studied theoretically in~\cite{fischer2010bounding}. 
%
There are a number of useful variants of the basic CD method. 
One can use $k$ Gibbs updates, instead of just one, in what is known as the CD$_k$ method. 
The larger $k$, the more one can expect the samples to follow the model distribution. 
In this spirit, there is also the persistent CD method (PCD)~\cite{Tieleman:2008:TRB:1390156.1390290}, where each sampling chain is initialized at previous samples, rather than at examples form the data set. 
Another useful technique in this context is parallel tempering~\cite{NIPS2009_3717, 10.1007/978-3-642-33275-3_2}. 
Moreover, basic gradient methods are often combined with other strategies, such as momentum, weight decay, pre-conditioners, second order methods. 
For more details see the introduction to training RBMs~\cite{fischer2014training} and the useful practical guide~\cite{Hinton2010}.

\paragraph{Natural gradient}

A natural modification of the standard gradient method is the \emph{natural gradient}, which is based on the notion that the parameter space has an underlying geometric structure. 
This is the point of view of~\emph{information geometry}~\cite{amari1985differential,amari2007methods,igaia}. 
A recent mathematical account on this topic is given in the book~\cite{InfGeo}. 
The natural gradient method was popularized with Amari's paper~\cite{Amari:1998:NGW:287476.287477}, which discusses how this method is efficient in learning. 
In this setting, the ordinary gradient is replaced by a Riemannian gradient, which leads to a parameter update rule of the form 
\begin{equation}
\theta^{(t+1)} = \theta^{(t)} + \alpha_t G^{-1}(\theta^{(t)})\nabla L(\theta^{(t)}), 
\end{equation}
where $G$ is the Fisher information~\cite{Rao45statmanifold}. 
For a given parametric model $\{p_\theta\colon \theta\in\mathbb{R}^d\}$, 
the Fisher information is defined as 
\begin{equation*}
G(\theta) = \mathbb{E}_{\theta}\big[ \nabla \log p(X;\theta) \cdot \nabla^\top \log p(X;\theta)  \big] .  
\end{equation*} 
Here $\mathbb{E}_\theta[\cdot]$ denotes expectation with respect to the model distribution $p(X;\theta)\equiv p_\theta$. 
Amari, Kurata, and Nagaoka~\cite{125867} discuss the statistical meaning of the Fisher metric. 
The inverse Fisher matrix divided by the number of observations describes the behavior of the expected square error (covariance matrix) of the maximum likelihood estimator. 

For an exponential family model with sufficient statistics $F\colon \Xcal\to \mathbb{R}^d$ and log-partition function $\psi(\theta) = \log Z(\theta)$, 
the Fisher matrix can be given as the Hessian of the log-partition function, as 
$$
G(\theta) = \nabla \nabla^\top \psi(\theta) = 
\mathbb{E}_\theta[F \cdot F^\top] - \mathbb{E}_\theta[F] \cdot \mathbb{E}_\theta[F]^\top = \operatorname{Cov}_{\theta}[F],
$$ 
which is the covariance of $F$ with respect to the exponential family distribution. 
This matrix is full rank iff the exponential family parametrization is minimal, meaning that the functions $F_1,\ldots, F_d\colon \Xcal\to\mathbb{R}$ are linearly independent and do not contain the constant function $1$ in their linear span.

Consider now the RBM model as the set of visible marginals of an exponential family with sufficient statistics $F\colon \Xcal\times \Ycal\mapsto \mathbb{R}^{d}$. 
The gradient of the visible log-probabilities is 
\begin{equation}
\nabla\log p(x;\theta) = \mathbb{E}_{\theta} [F|x] - \mathbb{E}_{\theta} [F], 
\end{equation}
where $\mathbb{E}_{\theta}[F|x] = \sum_{y} F(x,y) p(y|x;\theta)$ is the conditional expectation of $F$, given the visible state $x$, 
and $\mathbb{E}_{\theta} [F]=\sum_{x,y}F(x,y)p(x,y;\theta)$ is the expectation with respect to the joint distribution over visible and hidden states. 
The Fisher matrix takes the form 
\begin{eqnarray*}
G(\theta) 
&=& \mathbb{E}_{\theta} [\mathbb{E}_{\theta}[F|X] \cdot \mathbb{E}_{\theta}[F|X]^\top] - \mathbb{E}_\theta[F] \cdot \mathbb{E}_\theta[F]^\top \\
&=& 
\operatorname{Cov}_{\theta}[\mathbb{E}_{\theta}[F|X]]. 
\end{eqnarray*}
The rank of this matrix is equal to the rank of 
the Jacobian $J(\theta)= [\nabla p(x;\theta)]_x$ of the parametrization of the visible marginal distributions. 
Verifying whether and when the Fisher matrix of the RBM has full rank, is a non-trivial problem that we will discuss further in Section~\ref{sec:dimension}. 

In models with hidden variables, the Fisher matrix is not always full rank. 
An area that studies the statistical effects of this is \emph{singular learning theory}; see~\cite{Watanabe:2009:AGS:1655832,aoyagi:2010}. 
In practice, for the purpose of parameter optimization, the natural gradient works well even when the model involves singularities, at least so long as the parameter updates don't step into the singular set. 
The advantages of the natural gradient over the regular gradient have been demonstrated in numerous applications. 
It tends to be better at handling plateaus, thus reducing the number of required parameter updates, and also to find better local optimizers. 
On the other hand, computing the Fisher matrix and its inverse is challenging for large systems. 
Approximations of the relevant expectation values still require a computational overhead over the regular gradient, and in some cases, it is not clear how to balance optimization with other statistical considerations. 
Approximating the Fisher matrix in an efficient and effective way is an active topic of research. RBMs have been discussed specifically in~\cite{Pascanu+Bengio-ICLR2014,pmlr-v37-grosse15}. 
Following the notions of the natural gradient, recent works also investigate alternatives and variants of the Fisher metric, for instance based on the Wasserstein metric~\cite{Montavon:2016:WTR:3157382.3157513,wli}.

\paragraph{Doubly minimization, EM algorithm}

Amari~\cite[Section~8.1.3]{igaia} discusses an alternative view on the maximum likelihood estimation problem in probability models with hidden variables. 
See also~\cite{125867,csiszar_information_1984}. 
The idea is to regard this as an optimization problem over the model of joint distributions of both visible and hidden variables. 
Given an empirical data distribution $p_V$ over visible states $x\in\Xcal$, 
consider the set of joint distributions over $(x,y)\in\Xcal \times\Ycal$ that are compatible with $p_V$: 
\begin{equation*}
E = 
\Big\{  p(x, y) \colon \sum_{y\in\Ycal} p(x,y) = p_V(x) \Big\}. 
\end{equation*} 
This \emph{data manifold} $E$, being defined by linear equality constraints, is a special type of linear model. 
Note that it can be written as $E = \{p(x, y) = p_V(x)p(y|x) \}$, where we fix the marginal distribution $p_V(x)$ and are free to choose arbitrary conditional distributions $p(y|x)$ of hidden states given the visible states. 

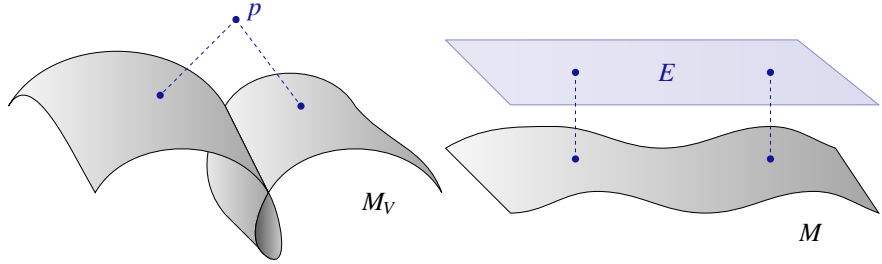
\begin{figure}
	\centering
	\scalebox{.48}{
		\begin{tikzpicture}[x=1.2cm,y=1.2cm]
		\coordinate (a) at (0,0);
		\coordinate (b) at (5,0);
		\coordinate (c) at (6,-2);
		\coordinate (d) at (2,-2);
		
		\coordinate (e) at (8,0);
		\coordinate (f) at (10,-2);
		
		\coordinate (g) at (5,-2.5);
		\coordinate (h) at (6,-3.5);
		\coordinate (h1) at (6,-3.5);
		
		\shade[left color=gray!10,right color=gray!70, draw, thick] 
		(b) --
		(c) to[out=240,in=150] 
		(h1) --
		(g) to[out=130,in=240] 
		(b); 
				
		\shade[left color=gray!10,right color=gray!70, draw, thick] 
		(b) to[out=60,in=120] (e) to[out=310,in=110] (f) to[out=120,in=60] (c) -- (b);
		
		\shade[left color=gray!5,right color=gray!70, draw, thick] 
		(a) to[out=60,in=120] (b) -- (c) to[out=120,in=60] (d) to[out=120,in=60] (a);

		\shade[left color=black!80,right color=gray!10, draw, thick] 
		(c) to[out=240,in=150] 
		(h) to[out=330,in=300] 
		(c);

		\coordinate (p) at (5.25,2);
		\coordinate (p1) at (3.5,.25);
		\coordinate (p2) at (6.75,0);
		
		\path[draw, thick, dashed, bl] (p) -- (p1);
		\path[draw, thick, dashed, bl] (p) -- (p2);
		
		\fill[bl] (p) circle (1mm)
		node [xshift=5mm,yshift=2mm] {\huge$p$};
		
		\fill[bl] (p1) circle (1mm)
		node [xshift=-5mm,yshift=0mm] {};
		
		\fill[bl] (p2) circle (1mm)
		node [xshift=5mm,yshift=0mm] {};
		
		\node at (9,-2.25) (obs) {\begin{minipage}{2cm}\huge$M_{V}$\end{minipage}};
		\end{tikzpicture} 
		\begin{tikzpicture}[x=1.2cm,y=1.2cm]	
		\coordinate (a) at (0,0.25);
		\coordinate (b) at (2.5,.75);
		\coordinate (c) at (5,0.25);
		\coordinate (d) at (7.5,.75);
		\coordinate (e) at (9,0.25);
		
		\coordinate (j) at (1.5,-1.25);
		\coordinate (i) at (3.5,-.75);
		\coordinate (h) at (6,-1.25);
		\coordinate (g) at (8.5,-.75);
		\coordinate (f) at (10,-1.25);

		\shade[left color=gray!5,right color=gray!70, draw, thick] 
		(a) to[out=30,in=180] 
		(b) to[out=0,in=180] 
		(c) to[out=0,in=180] 
		(d) to[out=0,in=160] 
		(e) -- 
		(f) to[out=160,in=0] 
		(g) to[out=180,in=0] 
		(h) to[out=180,in=0] 
		(i) to[out=180,in=0] 
		(j) --
		(a);
		
		\coordinate (k) at (0,2.75);
		\coordinate (l) at (8.125,2.75);
		\coordinate (m) at (10,1.25);
		\coordinate (n) at (1.5,1.25);
		
		\shade[left color=bl!10,right color=bl!15, draw, thick, color=bl!50] 
		(k) --
		(l) --
		(m) --
		(n) -- (k);
		
		\coordinate (p1) at (3,2);
		\coordinate (p2) at (7.5,2);
		\coordinate (q1) at (3,0);
		\coordinate (q2) at (7.5,0);
		
		\path[draw, thick, dashed, bl] (p1) -- (q1);
		\path[draw, thick, dashed, bl] (p2) -- (q2);
		
		\fill[bl] (p1) circle (1mm)
		node [xshift=-5mm,yshift=0mm] {}; 
		\fill[bl] (p2) circle (1mm)
		node [xshift=5mm,yshift=0mm] {}; 

		\fill[bl] (q1) circle (1mm)
		node [xshift=-5mm,yshift=-0mm] {};
		
		\fill[bl] (q2) circle (1mm)
		node [xshift=5mm,yshift=0mm] {};
		
		\node at (9,-1.7) (full) {\begin{minipage}{2cm}\huge$M$\end{minipage}};
		\node at (5.75,2) (datmf) {\begin{minipage}{2cm}\huge\textcolor{bl}{$E$}\end{minipage}};
		
		\node at (3,-2.5) (sp) {};
		\end{tikzpicture}
	}
	\caption{Schematic illustration of the maximum likelihood estimation problem over the set of visible marginal distributions $M_V$, and over the set of joint distributions $M$ prior to marginalization. }
	\label{fig:jointvis}
\end{figure}

Taking this view, we no longer minimize the divergence from $p_V$ to our model $M_V$ of visible marginal distributions $q_V(x;\theta) = \sum_{y} q(x, y;\theta)$, 
but rather we seek for the distributions $q(x, y;\theta)$ in the model $M$ of joint distributions, with the smallest divergence from the data manifold $E$. 
The situation is illustrated schematically in Figure~\ref{fig:jointvis}. 

When working with the data manifold $E$ and the joint model $M$, the maximum likelihood estimation problem becomes a double minimization problem 
\begin{equation}
\min_{p\in E ,  q\in M} D(p\| q). 
\label{eq:doubl}
\end{equation}
The minimum of this problem equals the minimum of the original problem 
$$
\min_{q_V\in M_V}D(p_V \| q_V).
$$ 
To see this, use the chain rule for probability, $P(x,y)=P(x)P(y|x)$, to write
\begin{align*}
\min_{p\in E, q\in M} D(p\|q) 
 =&  \min_{p\in E, q\in M}\sum_x\sum_y  p(x,y) \log \frac{p(x,y)}{q(x,y)}\\
=& \min_{q\in M} \sum_x p_V(x) \log \frac{p_V(x)}{q_V(x)} + \min_{p(y|x)}\sum_x p_V(x)\sum_y p(y|x)\log \frac{p(y|x) }{q(y|x)}\\
=& \min_{q_V\in M_V} D(p_V\|q_V). 
\end{align*}
For simplicity of exposition, we are assuming that the sets $E$ and $M$ are so that the minimum can be attained, e.g., they are closed. 

The expression~\eqref{eq:doubl} hints at an approach to computing the minimizers. 
Namely, we can iteratively minimize with respect to each of the two arguments. 
\begin{itemize}
\item 
For any fixed value of the second argument, $q\in M$, minimization of the divergence over the first argument $p\in E$ is a convex problem, because $E$ is a linear model. 
This is solved by the \emph{$e$-projection} of $q$ onto $E$, which is given simply by setting $p(y|x)=q(y|x)$. 
\item 
For any fixed value of the first argument, $p\in E$, the minimization over the second argument $q\in M$ is also a convex problem, because $M$ is an exponential family. 
It is solved by the \emph{$m$-projection} of $p$ onto $M$, which is given by the unique distribution $q$ in $M$ for which $\sum_{x,y}F(x,y)q(x,y) = \sum_{x,y}F(x,y)p(x,y)$. 
\end{itemize}
This procedure corresponds to the expectation maximization (EM) algorithm~\cite{10.2307/2984875}.

\paragraph{Optimization landscape}

In general, for a model with hidden variables, we must assume that the log-likelihood function $L(\theta)$ is non-concave. 
Gradient methods and other local techniques, such as contrastive divergence and EM, may only allow us to reach critical points or locally optimal solutions. 
The structure of the optimization landscape and critical points of these methods is the subject of current studies. 
In Section~\ref{sec:semialgebraic} we discuss results from~\cite{SeigalMontufar} showing that an RBM model can indeed have several local optimizers with different values of the likelihood function, but also that in some cases, the optimization problem may be solvable in closed form. 
%

\section{Dimension}
\label{sec:dimension}

From a geometric standpoint, a basic question we are interested in, is the dimension of the set of distributions that can be represented by our probability model. 
The dimension is useful when comparing a model against other models, or when testing hypotheses expressed in terms of equality constraints. 
Under mild conditions, if the dimension is equal to the number of parameters, then the Fisher matrix is regular almost everywhere and the model is generically locally identifiable. 

A Boltzmann machine with all units observed is an exponential family, and its dimension can be calculated simply as the dimension of the linear space spanned by the sufficient statistics, disregarding constant functions. 
This is precisely equal to the number of parameters of the model, since the statistics associated with each of the parameters, bias and interaction weights, are linearly independent. 

When some of the units of the Boltzmann machine are hidden, as is usually the case, the set of observable distributions is no longer an exponential family,  but rather a linear projection of an exponential family. 
The marginalization map takes the high dimensional simplex $\Delta_{\Xcal\times\Ycal}$ to the low dimensional simplex $\Delta_{\Xcal}$. Such a projection can in principle collapse the dimension of the set that is being projected. 
A simple example where this happens is the set of product distributions. 
The visible marginals of an independence model are simply the independent distributions of the observed variables, 
meaning that the hidden variables and their parameters do not contribute to the dimension of the observable model. 
Another well-known example is the set of mixtures of three product distributions of four binary variables. 
This model has dimension $13$, instead of $14$ that one would expect from the number of model parameters. 
Computing the dimension of probability models with hidden variables often corresponds to challenging problems in algebraic geometry, most prominently the dimension of secant varieties, which correspond to mixture models.

\paragraph{Tropical approach}

The first investigation of the dimension of the RBM model was by Cueto, Morton, and Sturmfels~\cite{Cueto2010}, using tools from tropical geometry and secant varieties. 
The tropical approach to the dimension of secant varieties was proposed by Draisma~\cite{Draisma}. 
It can be used in great generality, and it was also used to study non-binary versions of the RBM~\cite{montufar2013discrete}. 

As mentioned in Section~\ref{sec:definitions}, the tropical RBM consists of piecewise linear approximation of the log-probability vectors of the RBM. 
The dimension of the tropical RBM is often easy to estimate by combinatorial arguments. 
A theorem by Bieri and Groves~\cite{Bieri1984,Draisma} implies that the dimension of the tropical RBM model is a lower bound on the dimension of the original RBM model. 
Using this method,~\cite{Cueto2010} proved that the RBM model has the expected dimension for most combinations of $n$ and $m$. 
However, a number of cases were left open. In fact, for the tropical RBM those cases are still open. 
A different approach to the dimension of RBMs was proposed in~\cite{montufar2015dimension}, which allowed verifying the conjecture that it always has the expected dimension. In the following we discuss this approach and how it compares to the tropical approach. 

\paragraph{Jacobian rank of RBMs and mixtures of products}

The dimension of a smoothly parametrized model can be computed as the maximum rank of the Jacobian of the parametrization. 
For a parametrization $p(x;\theta) = \sum_y p(x,y;\theta)$, with $p(x,y;\theta) = \frac{1}{Z(\theta)} \exp ( \sum_{i} \theta^\top F(x,y))$, the columns of the Jacobian matrix are 
\begin{equation}
J_{: x}(\theta) = \sum_y p(x,y;\theta) ( F(x,y) - \sum_{x',y'} p(x',y';\theta)F(x',y')),\quad x\in\Xcal. 
\label{eq:jacobian}
\end{equation}
Now we need to consider the specific $F$ and evaluate the maximum rank of the matrix~$J$ over the parameter space. 
In order to simplify this, one possibility is to consider the limit of large parameters $\theta$. 
The corresponding limit distributions usually have a reduced support and the sum in~\eqref{eq:jacobian} has fewer nonzero terms. 
As shown in~\cite{montufar2015dimension}, the dimension bounds from the tropical approach can be obtained in this manner. 
On the other hand, it is clear that after taking such limits, it is only possible to lower bound the maximum rank. 
Another problem is that, when the number of parameters is close to the cardinality of $\Xcal$, the rank of the limit matrices is not always easy to compute, with block structure arguments leading to challenging combinatorial problems, such as accurately estimating the maximum cardinality of error correcting codes. 

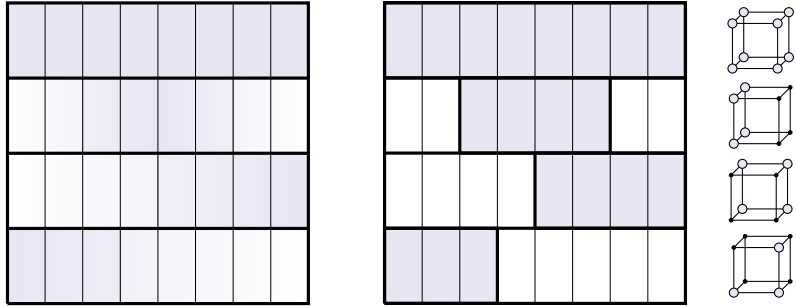
\begin{figure}
	\centering
\begin{tikzpicture}
\draw[ fill=bl!10] (0,3) -- (4,3) -- (4,4) -- (0,4) -- (0,3);
\shade[left color=white, right color=bl!10, opacity=1, very thin] (0,2) rectangle (2,3);
\shade[left color=bl!10, right color=white, opacity=1, very thin] (2,2) rectangle (4,3);
\draw[fill=bl!10, opacity=.3] (1,2) -- (3,2) -- (3,3) -- (1,3) -- (1,2);
\shade[left color=white,right color=bl!10, opacity=1, very thin] (0,1) rectangle (4,2);
\draw[fill=bl!10, opacity=.3, very thin] (2,1) -- (4,1) -- (4,2) -- (2,2) -- (2,1);
\shade[left color=bl!10,right color=white, opacity=1, very thin] (0,0) rectangle (4,1);
\draw[fill=bl!10, opacity=.3, very thin] (0,0) -- (1.5,0) -- (1.5,1) -- (0,1) -- (0,0);
\foreach \y in {1, 2, 3} {
	\draw[very thick] (0,\y) -- (4,\y);
}
\foreach \x in {1,2,...,8}{
	\draw[very thin] (.5*\x,0) -- (.5*\x,4);
}
\draw[very thick] (0,0) -- (4,0) -- (4,4) -- (0,4) -- (0,0);
\end{tikzpicture}
\qquad \quad
\begin{tikzpicture}
\draw[very thick, fill=bl!10] (0,3) -- (4,3) -- (4,4) -- (0,4) -- (0,3);
\draw[very thick, fill=bl!10] (1,2) -- (3,2) -- (3,3) -- (1,3) -- (1,2);
\draw[very thick, fill=bl!10] (2,1) -- (4,1) -- (4,2) -- (2,2) -- (2,1);
\draw[very thick, fill=bl!10] (0,0) -- (1.5,0) -- (1.5,1) -- (0,1) -- (0,0);
\foreach \x in {1,2,...,8}{
	\draw[very thin] (.5*\x,0) -- (.5*\x,4);
}
\draw[very thick] (0,0) -- (4,0) -- (4,4) -- (0,4) -- (0,0);
\foreach \y in {1, 2, 3} {
	\draw[very thick] (0,\y) -- (4,\y);
}
\node at (5,3.5) {\begin{tikzpicture}[x=.6cm,y=.6cm]
	\draw[thin] (0,0) --(1,0) -- (1,1) -- (0,1) -- (0,0);
	\draw[thin] (0.25,0.25) --(1.25,0.25) -- (1.25,1.25) -- (0.25,1.25) -- (0.25,0.25);
	\draw[thin] (0,0) -- (0.25,0.25);
	\draw[thin] (1,0) -- (1.25,0.25);
	\draw[thin] (0,1) -- (0.25,1.25);
	\draw[thin] (1,1) -- (1.25,1.25);
	
	\draw [fill=bl!10] (0,0) circle [radius=.1];
	\draw [fill=bl!10] (0,1) circle [radius=.1];
	\draw [fill=bl!10] (1,0) circle [radius=.1];
	\draw [fill=bl!10] (1,1) circle [radius=.1];
	
	\draw [fill=bl!10] (0.25,0.25) circle [radius=.1];
	\draw [fill=bl!10] (0.25,1.25) circle [radius=.1];
	\draw [fill=bl!10] (1.25,0.25) circle [radius=.1];
	\draw [fill=bl!10] (1.25,1.25) circle [radius=.1];
	\end{tikzpicture}
};
\node at (5,2.5) {\begin{tikzpicture}[x=.6cm,y=.6cm]
	\draw[thin] (0,0) --(1,0) -- (1,1) -- (0,1) -- (0,0);
	\draw[thin] (0.25,0.25) --(1.25,0.25) -- (1.25,1.25) -- (0.25,1.25) -- (0.25,0.25);
	\draw[thin] (0,0) -- (0.25,0.25);
	\draw[thin] (1,0) -- (1.25,0.25);
	\draw[thin] (0,1) -- (0.25,1.25);
	\draw[thin] (1,1) -- (1.25,1.25);
	
	\draw [fill=bl!10] (0,0) circle [radius=.1];
	\draw [fill=bl!10] (0,1) circle [radius=.1];
	\draw [fill] (1,0) circle [radius=.04];
	\draw [fill] (1,1) circle [radius=.04];
	
	\draw [fill=bl!10] (0.25,0.25) circle [radius=.1];
	\draw [fill=bl!10] (0.25,1.25) circle [radius=.1];
	\draw [fill] (1.25,0.25) circle [radius=.04];
	\draw [fill] (1.25,1.25) circle [radius=.04];
	\end{tikzpicture}
};
\node at (5, 1.5) {\begin{tikzpicture}[x=.6cm,y=.6cm]
	\draw[thin] (0,0) --(1,0) -- (1,1) -- (0,1) -- (0,0);
	\draw[thin] (0.25,0.25) --(1.25,0.25) -- (1.25,1.25) -- (0.25,1.25) -- (0.25,0.25);
	\draw[thin] (0,0) -- (0.25,0.25);
	\draw[thin] (1,0) -- (1.25,0.25);
	\draw[thin] (0,1) -- (0.25,1.25);
	\draw[thin] (1,1) -- (1.25,1.25);
	
	\draw [fill] (0,0) circle [radius=.04];
	\draw [fill] (0,1) circle [radius=.04];
	\draw [fill] (1,0) circle [radius=.04];
	\draw [fill] (1,1) circle [radius=.04];
	
	\draw [fill=bl!10] (0.25,0.25) circle [radius=.1];
	\draw [fill=bl!10] (0.25,1.25) circle [radius=.1];
	\draw [fill=bl!10] (1.25,0.25) circle [radius=.1];
	\draw [fill=bl!10] (1.25,1.25) circle [radius=.1];
	\end{tikzpicture}
};
\node at (5,.5) {\begin{tikzpicture}[x=.6cm,y=.6cm]
	\draw[thin] (0,0) --(1,0) -- (1,1) -- (0,1) -- (0,0);
	\draw[thin] (0.25,0.25) --(1.25,0.25) -- (1.25,1.25) -- (0.25,1.25) -- (0.25,0.25);
	\draw[thin] (0,0) -- (0.25,0.25);
	\draw[thin] (1,0) -- (1.25,0.25);
	\draw[thin] (0,1) -- (0.25,1.25);
	\draw[thin] (1,1) -- (1.25,1.25);
	
	\draw [fill=bl!10] (0,0) circle [radius=.1];
	\draw [fill] (0,1) circle [radius=.04];
	\draw [fill=bl!10] (1,0) circle [radius=.1];
	\draw [fill=bl!10] (1,1) circle [radius=.1];
	
	\draw [fill] (0.25,0.25) circle [radius=.04];
	\draw [fill] (0.25,1.25) circle [radius=.04];
	\draw [fill] (1.25,0.25) circle [radius=.04];
	\draw [fill] (1.25,1.25) circle [radius=.04];
	\end{tikzpicture}
};
\end{tikzpicture}
\caption{Illustration of the Jacobian matrix for an RBM with three visible and three hidden units, and its tropical counterpart, together with the corresponding slicings of the visible sufficient statistics. 
Rows correspond to model parameters and columns to visible states. }
\label{fig:jacob}
\end{figure}

For the analysis it is convenient to work with the denormalized model, which includes all positive scalar multiples of the probability distributions. The dimension of the original model is simply one less. 
Following~\eqref{eq:jacobian}, and as discussed in~\cite{montufar2015dimension}, the Jacobian for the denormalized RBM 
is equivalent to the matrix with columns 
\begin{equation}
\sum_y p(y|x;\theta) F(x,y) 
= \sum_y p(y|x;\theta) \hat x \otimes \hat y = \hat x \otimes \hat \sigma(W x + c), \quad x\in \Xcal, 
\label{eq:dimjac}
\end{equation}
where we write $\hat v = (1,v^\top)^\top$ for the vector $v$ with an additional~$1$. 
Here $\sigma(\cdot) = \exp(\cdot)/(1+\exp(\cdot))$ can be regarded as the derivative of the soft-plus function $\log(1+\exp(\cdot))$. 
The $j$th coordinate of $\sigma(Wx+c)$ ranges between $0$ and $1$, taking larger values the farther $x$ lies in the positive side of the hyperplane $H_j = \{r\in\mathbb{R}^V \colon W_{j:} r + c_j = 0\}$. 
In the case of the tropical RBM, the Jacobian is equivalent to the matrix with columns 
$$
\hat x \otimes \hat {\mathds{1}}_{[W x +c]_+},\quad x\in\Xcal,
$$ 
where now $\mathds{1}_{[\cdot]_+}$ corresponds to the derivative of the rectification non-linearity $[\cdot]_+$. 
The $j$th coordinate indicates whether the point $x$ lies on the positive side of the hyperplane $H_j$. 
%
The matrices for the RBM and the tropical RBM are illustrated in Figure~\ref{fig:jacob}. 

In~\cite{montufar2015dimension} it is shown that~\eqref{eq:dimjac} can approximate the following matrix, equivalent to the Jacobian of a mixture of $m+1$ product distributions model, arbitrarily well at generic parameters:  
$$
\hat x\otimes \hat{\sigma'}(\tilde Wx + \tilde c),\quad x\in \Xcal. 
$$ 
Here $\sigma'(\tilde W x+\tilde c) =\frac{\exp(\tilde W x+\tilde c)}{\sum_j \exp(\tilde W_{j:}x+\tilde c_j)}$ is a soft-max unit. 
In turn, the dimension of the RBM model is bounded below by the dimension of the mixture model. 
But the results from~\cite{Catalisano2011} imply that mixture models of binary product distributions have the expected dimension (except in one case, which for the RBM can be verified by other means). 
This implies that the RBM model always has the expected dimension:  

\begin{theorem}[{\cite[Corollary~26]{montufar2015dimension}}]
	For any $n,m\in\mathbb{N}$ the model $\RBM_{n,m}$, with $n$ visible and $m$ hidden binary units, has dimension $\min\{2^n-1, (n+1)(m+1)-1\}$. 
\end{theorem}
This result implies that, unless the number of parameters exceeds $2^n-1$, almost every probability distribution in the RBM model can be represented only by finitely many different choices of the parameters. 
One trivial way in which the parameters are not unique, is that we can permute the hidden units without changing the represented distributions, $\sum_{j\in H} \log(1+\exp (w_{j} x + c_{j})) = \sum_{j\in H}\log(1+\exp (w_{\pi(j)} x + c_{\pi(j)}))$ for all $\pi\in H!$.  
On the other hand, there are also some few probability distributions that can be represented by infinitely many different choices of the parameters. 
For instance, if $w_j=0$, then the choice of $c_j$ is immaterial. 

The characterization of the parameter fibers $\{\theta\in\mathbb{R}^d\colon p_\theta=p\}$ of the distributions $p$ that can be represented by an RBM model is an important problem, with implications on the parameter optimization problem, which still requires more investigation. 
We can ask in the first place whether a given distribution $p$ can be represented by an RBM model. We discuss this in the next section.

\section{Representational power}
\label{sec:representational}

The representational power of a probability model can be studied from various angles. 
An idea is that each parameter allows to model certain features or properties of the probability distributions. The question then is how to describe and interpret these features. 
As we have seen, each hidden unit of an RBM can be interpreted as contributing entrywise multiplicative factors which are arbitrary mixtures of two product distributions. 
Alternatively, each hidden unit can be interpreted as adding a soft-plus unit to the negative energy function of the visible distributions. 

Now we want to relate these degrees of freedom with the degrees of freedom of other families of distributions for which we have a good intuition, or for which we can maximize the likelihood function in closed form and compute metrics of the representational power, such as the maximum divergence. 
The natural approach to this problem is by showing that there exist choices of parameters for which the model realizes a given distribution of interest, or, more generally, a class of distributions of interest. 
We note that another approach, which we will discuss in Section~\ref{sec:semialgebraic}, is by showing that any constraints that apply on the set of distributions from the RBM are less stringent than the constraints that apply on the distributions of interest.

\paragraph{Overview} 
\label{section:topRBM}

The representational power of RBMs has been studied in many works. 
Le~Roux and Bengio~\cite{LeRoux:2008:RPR:1374176.1374187} showed that each hidden unit of an RBM can model the probability of one elementary event.  Freund and Haussler~\cite{Freund:1994:ULD:902676} used similar arguments to discuss universal approximation. 
In~\cite{montufar2011refinements} it was shown that each hidden unit can model the probability of two elementary events of Hamming distance one, which implied improved bounds on the minimal number of hidden units that is sufficient for universal approximation. 
Generalizing this,~\cite{NIPS2011_0307} showed that each hidden unit can model a block of elementary events with a weighted product distribution, provided certain conditions on the support sets are satisfied. 
Another line of ideas was due to~\cite{Younes1996109}, showing that each hidden unit can model the coefficient of a monomial in a polynomial representation of the energy function. 
This analysis was refined in~\cite{montufar2016hierarchical}, showing that each hidden unit can model the coefficients of as many as $n$ monomials in the energy function. 

\paragraph{Mixtures of products and partition models}


We discuss a result from~\cite{NIPS2011_0307} showing that an RBM with $m$ hidden units can represent mixtures of $m+1$ product distributions, provided the support sets of $m$ of the mixture components are disjoint. 
The support of a distribution $p$ on $\Xcal$ is $\supp(p):=\{x\in\Xcal\colon p(x)>0\}$. 
The idea is as follows. 
Consider an entrywise product of the form 
\begin{equation}
p_0(x) (1 + \lambda p_1(x)) = p_0(x) + \lambda p_0(x)p_1(x), \quad x\in\Xcal. 
\label{eq:ewpr}
\end{equation}
If $p_0$ and $p_1$ are product distributions, then so is $p_2=p_0p_1$. 
This is a direct consequence of the fact that the set of product distributions has an affine set of exponential parameters, $\exp(w_0^\top x)\exp(w_1^\top x)=\exp((w_0+w_1)^\top x)=\exp(w_2^\top x)$. 
In turn, an entrywise product of the form~\eqref{eq:ewpr} expresses a linear combination of product distributions, provided that $p_0$ and $p_1$ are product distributions. 
The last requirement can be relaxed to hold only over the intersection of the support sets of $p_0$ and $p_1$, since the entrywise product will vanish on the other entries either way. 
When we renormalize, the linear combination becomes a mixture of product distributions, whereby the relative mixture weights are controlled by $\lambda$.

Now recall from Section~\ref{sec:definitions} that the RBM distributions can be written as 
\begin{equation}
p(x;\theta) = \frac{1}{Z(\theta)}\exp(b^\top x) \prod_{j\in H}(1+ \exp(c_j)\exp(W_{j:}x )).  
\label{eq:rbmdsf}
\end{equation} 
By the previous discussion, we can interpret each factor in~\eqref{eq:rbmdsf}  as adding a mixture component $p_j(x) = \frac{1}{Z}\exp(W_{j:}x )$, which is a product distribution, so long as the distribution obtained from the preceding factors is a product distribution over the support of $p_j$. 
Being an exponential family distribution, $p_j$ has full support, but it can approximate product distributions with restricted support arbitrarily well.

A similar discussion applies to non-binary variables, as shown in~\cite{montufar2013discrete}.  
We denote by $\RBM_{\Xcal,\Ycal}$ the RBM with visible states $\Xcal=\Xcal_1\times\cdots\times\Xcal_n$ and hidden states $\Ycal=\Ycal_1\times\cdots\times\Ycal_m$. This is the set of marginals of the exponential family with sufficient statistics given by the Kronecker product of the statistics of the independence models on $\Xcal$ and $\Ycal$, respectively. 

\begin{theorem}[{\cite[Theorem~3]{montufar2014universal}}]
	\label{theorem:RBMs}
	Let $\Xcal=\Xcal_1\times\cdots\times \Xcal_n$ and $\Ycal=\Ycal_1\times\cdots\times \Ycal_m$ be finite sets. 
	The model $\RBM_{\Xcal,\Ycal}$ can approximate any mixture distribution $p(x)=\sum_{i=0}^m \lambda_i p_i(x)$, $x\in\Xcal$, arbitrarily well, where $p_0$ is any product distribution, and $p_i$ are respectively for all $i\in[m]$, any mixtures of $(|\Ycal_i|-1)$ product distributions, with support sets satisfying $\supp(p_i)\cap\supp(p_j)=\emptyset$ for all $1\leq i < j\leq m$. 
\end{theorem}
In particular, the binary $\RBM_{n,m}$ can approximate, to within any desired degree of accuracy, any mixture of $m+1$ product distributions with disjoint supports. 
Given a collection of disjoint sets $A_1,\ldots, A_{m+1}\subseteq \Xcal$, 
the set of mixtures $p = \sum_j \lambda_j p_j$, where each $p_j$ is a product distribution with support set $A_j$, is an exponential family on $\cup_j A_j$. More precisely, its topological closure coincides with that of an exponential family with sufficient statistics $\mathds{1}_{A_j}$, $\mathds{1}_{A_j}x_i$, $i= 1,\ldots,n$, $j=1,\ldots, m+1$. 
Theorem~\ref{theorem:RBMs} shows that an RBM can represent all such exponential families, for all choices of disjoint sets $A_1,\ldots, A_{m+1}$. 

A \emph{partition model} is a special type of mixture model, consisting of all mixtures of a fixed set of uniform distributions on disjoint support sets. 
Partition models are interesting not only because of their simplicity, but also because they are optimally approximating exponential families of a given dimension. If all support sets of the components, or blocks, have the same size, then the partition model attains the smallest uniform approximation error, measured in terms of the Kullback-Leibler divergence, among all exponential families that have the same dimension~\cite{Rauh13:Optimal_Expfams}. 
The previous theorem shows that RBMs can approximate certain partition models arbitrarily well. In particular we have: 

\begin{corollary}
	\label{lemma:distrRBMs}
	Let $\Xcal=\Xcal_1\times\cdots\times \Xcal_n$ and $\Ycal=\Ycal_1\times\cdots\times \Ycal_m$ be finite sets. 	
	Let $\Pcal$ be the partition model with partition blocks  $\{x_1\}\times\cdots\times\{x_k\}\times\Xcal_{k+1}\times\cdots\times\Xcal_n$ for all $(x_1,\ldots,x_k)\in\Xcal_{1}\times\cdots\times\Xcal_k$. If $ 1+\sum_{j\in[m]}(|\Ycal_j|-1) \geq (\prod_{i\in [k]}|\Xcal_i| ) / \max_{j\in[k]}|\Xcal_j|$, then each distribution contained in $\Pcal$ can be approximated arbitrarily well by distributions from $\RBM_{\Xcal,\Ycal}$. 
\end{corollary}

\paragraph{Hierarchical models}
\label{sec:hierarchicalmodels}
Intuitively, each hidden unit of an RBM should be able to mediate certain interactions between the visible units. 
To make this more concrete, we may ask which distributions from a hierarchical model can be expressed in terms of an RBM, or which parameters of a hierarchical model can be modeled in terms of the hidden units of an RBM. 
Younes~\cite{Younes1996109} showed that a binary hierarchical model with a total of $K$ pure higher order interactions can be modeled by an RBM with $K$ hidden units. 
Later,~\cite{montufar2016hierarchical} showed that each hidden unit of an RBM can model several parameters of a hierarchical model simultaneously.

Consider a set $S\subseteq2^V$ of subsets of $V$. 
A hierarchical model with interactions $S$ is defined as the set of probability distributions $p$ that can be factorized as 
\begin{equation}
p(x) = \prod_{\lambda\in S} \psi_\lambda(x), \quad x\in \Xcal, 
\label{eq:factorizationhierarchical}
\end{equation}
where each $\psi_\lambda\colon \Xcal\to \mathbb{R}_+$ is a positive valued function that only depends on the coordinates $\lambda$, i.e., satisfies $\psi_\lambda(x)=\psi_\lambda(x')$ whenever $x_i=x'_i$ for all $i\in\lambda$. 
In practice, we choose a basis to express the potentials as parametrized functions. 
The set $S$ is conveniently defined as the set of cliques of a graph $G=(V,E)$, and hence these models are also known as hierarchical graphical models. 
These models are very intuitive and have been studied in great detail. 
Each factor $\psi_\lambda$ is interpreted as allowing us to model arbitrary interactions between the variables $x_i$, $i\in\lambda$, independently of the variables $x_j$, $j\in V\setminus\lambda$. 
Hence, they are a good reference to compare the representational power other models, which is what we want to do for RBMs in the following.

\begin{figure}
	\centering
	\includegraphics[scale=.9]{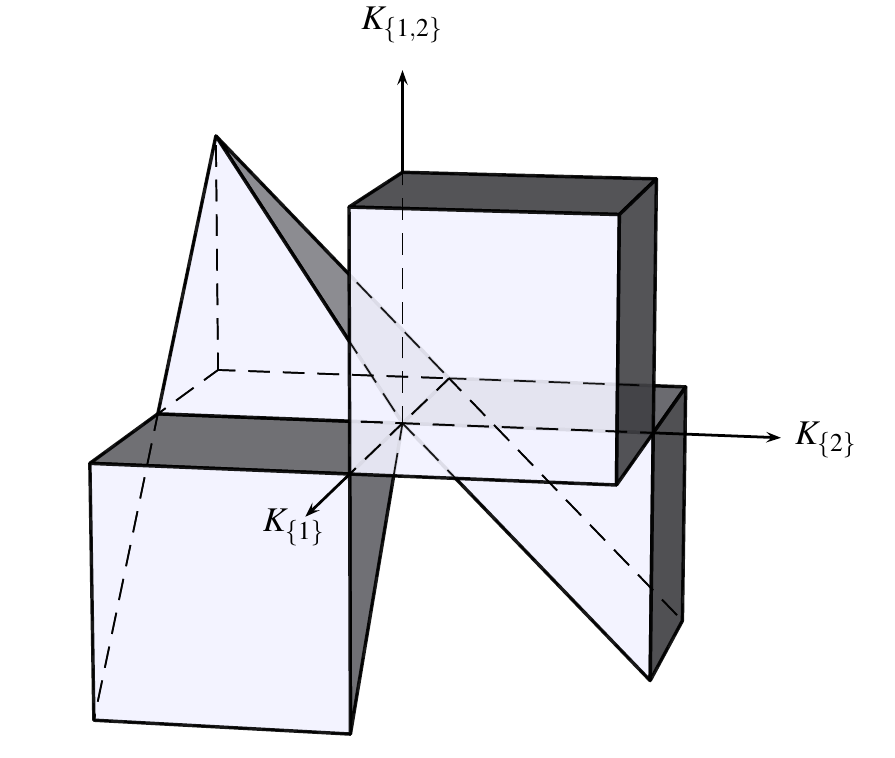}
	\caption{Interaction coefficients expressible by one RBM hidden unit. 
		Shown is the set of coefficients $(K_{\{1\}},K_{\{2\}},K_{\{1,2\}})\in\mathbb{R}^3$, clipped to a cube centered at the origin, of the polynomials $K_\emptyset + K_{\{1\}}x_1 + K_{\{2\}}x_2 + K_{\{1,2\}}x_1 x_2$ expressible in terms of a soft-plus unit on binary inputs. 
		Figure adapted from~\cite{montufar2016hierarchical}. 
	}
	\label{fig:softplus}
\end{figure}

At a high level, the difficulty of comparing RBMs and hierarchical models stems from the fact that their parameters contribute different types of degrees of freedom. 
While a hidden unit can implement interactions among all visible units it is connected to, certain constraints apply on the values of these interactions.  
For example, the set of interaction coefficients among two visible variables that can be modeled by one hidden unit is shown in Figure~\ref{fig:softplus}. 

To proceed with more details, we first fix a coordinate system. 
Hierarchical models are conveniently expressed in terms of a basis of orthogonal functions known as characters. 
For each $\lambda\subseteq V$ we have a function 
\begin{equation*}
\sigma_\lambda(x)= \prod_{i\in\lambda} (-1)^{x_i}, \quad x\in\{0,1\}^V.  
\end{equation*}
The functions $\sigma_{\lambda}$, $\lambda\subseteq V$, are orthogonal, with $\sum_x \sigma_{\lambda}(x) \sigma_\mu(x)=2^n \delta_{\lambda,\mu}$.  
In turn, we can express any given vector $l\in\mathbb{R}^{\{0,1\}^V}$ as 
\begin{equation*}
l(x) = \sum_{\lambda\subseteq V} J_\lambda \sigma_\lambda(x), \quad x\in\{0,1\}^V, 
\end{equation*}
where the coefficients are given by 
\begin{equation*}
J_\lambda = \frac{1}{2^n}\sum_{x\in\{0,1\}^V} \sigma_\lambda(x) l(x), \quad \lambda\subseteq V. 
\end{equation*}
The change of coordinates from the standard basis $\delta_x$, $x\in\{0,1\}^V$, to the basis of characters $\sigma_\lambda$, $\lambda\subseteq V$, can be interpreted as a M\"obius inversion, or also as a Fourier transform. 

If we replaced the states $\{0,1\}$ with $\{+1,-1\}$, we could write each $\sigma_{\lambda}$ as a monomial $\prod_{i\in\lambda}x_i$. 
%
But we can also use a basis of monomials without changing the states. 
For each $\lambda\subseteq V$, let 
\begin{equation}
\pi_\lambda(x) = \prod_{i\in \lambda} x_i , \quad x\in\{0,1\}^V. 
\end{equation}
Although this is no longer an orthogonal basis, it is conceptually simple and very frequently used in practice. 
Moreover, for an inclusion closed set $S\subseteq 2^V$, the span of $\pi_\lambda$, $\lambda \in S$, equals that of $\sigma_{\lambda}$, $\lambda \in S$, 
such that both bases have the same hierarchical coordinate sub-spaces.

For an inclusion closed set $S\subseteq 2^V$, the binary hierarchical model with interactions $S$ can be parametrized as the exponential family $\Ecal_S$ of distributions of the form 
\begin{equation}
p(x) = \frac{1}{Z}\exp\Big(\sum_{\lambda\in S} J_\lambda \prod_{i\in\lambda} x_i\Big), \quad x\in \{0,1\}^V, 
\label{eq:hierarchical}
\end{equation}
with parameters $J_\lambda\in\mathbb{R}$, $\lambda\in S$.

Now we proceed with the representation of the parameters of a hierarchical model in terms of an RBM. 
Recall that the log-probabilities $l=\log(p)$ in the model $\RBM_{n,m}$ are sums of a linear unit and $m$ soft-plus units. 
For a linear unit $w^\top x +c$, the polynomial coefficients are simply $K_\emptyset = c$, $K_{\{i\}} = w_i$, $i\in V$, and $K_\lambda =0$ for all $\lambda\subseteq V$ with $|\lambda|\geq 2$. 
For a soft-plus unit,~\cite{montufar2016hierarchical} obtains a partial characterization of the possible polynomial coefficients. In particular, it shows the following.  

\begin{lemma}[{\cite[Lemma~5]{montufar2016hierarchical}}]
	\label{generallemma}
	Consider a subset $B\subseteq V$, and let $J_{B\cup\{j\}}\in\mathbb{R}$, $j\in V\setminus B$, and $\epsilon>0$. 
	Then there are $w\in\mathbb{R}^{V}$ and $c\in\mathbb{R}$ such that the soft-plus unit $\log(1 + \exp(w^\top  x  + c))$ is equal to a polynomial $\sum_\lambda K_\lambda \prod_{i\in\lambda}x_i$ with coefficients satisfying 
	$|K_{B\cup\{j\}} - J_{B\cup\{j\}} |\leq \epsilon$ for all $j\in V\setminus B$, and $|K_C|\leq \epsilon$ for all $C\neq B, B \cup\{j\}$, $j\in V\setminus B$. 
\end{lemma}
This says that each hidden unit of an RBM can model arbitrarily the parameters of a hierarchical model corresponding to the monomials that cover $\prod_{i\in B}x_i$, for any fixed choice of $B\subseteq V$, while at the same time setting all other parameters arbitrarily close to zero, except for the parameter associated with $\prod_{i\in B}x_i$, whose value may be coupled to the values of the other parameters. 

We can use this result to describe hierarchical models that can be represented by an RBM. 
Since each hidden unit of the RBM can model certain subsets of parameters of hierarchical models, 
we just need to find a sufficiently large number of hidden units which together can model all the required parameters. 
For example: 

\begin{itemize}
	\item $\RBM_{3,1}$ contains the hierarchical models 
	$\Ecal_S$ with 
	$S=\{\{1\},\{2\},\{3\},\{1,2\},\{1,3\}\}$, 
	$S=\{\{1\},\{2\},\{3\},\{1,2\},\{2,3\}\}$, 
	$S=\{\{1\},\{2\},\{3\},\{1,3\},\{2,3\}\}$. 
	It does not contain the \emph{no-three-way interaction model},  
	with $S=S_2=\{\{1\},\{2\},\{3\},\{1,2\},\{1,3\},\{2,3\}\}$. 
	\item 
	The model $\RBM_{3,2}$ contains the no-three-way interaction model  
	$\Ecal_S$ with $S=S_2$. 
	It does not contain the full interaction model, with $S=S_3$. 
	In particular, this model is not a universal approximator. 
\end{itemize}

In general, finding a minimal cover of the relevant set of parameters of hierarchical models in terms of subsets of parameters of the form described in Lemma~\ref{generallemma} relates to well-known problems in the theory of combinatorial designs. 
For $S$ consisting of all sets up to a given cardinality, we can obtain the following bounds. 

\begin{theorem}[{\cite[Theorem~11]{montufar2016hierarchical}}]
	\label{theorem2a}
	Let $1\leq k\leq n$ and $\Xcal = \{0,1\}^V$. 
	Every distribution from the hierarchical model $\Ecal_{S_k}$, with $S_k=\{ \lambda\subseteq V \colon |\lambda|\leq k\}$,  
	can be approximated arbitrarily well by distributions from $\RBM_{n,m}$ whenever 
	$$m\geq \min \Big\{ \sum_{j=2}^k \binom{n-1}{j-1}, \frac{\log(n-1)+1}{n+1}\sum_{j=2}^k\binom{n+1}{j}\Big\}.$$ 
\end{theorem}
We note that in specific cases there are sharper bounds available, listed in~\cite{montufar2016hierarchical}. 

The hidden units and parameters of an RBM can be employed to model different kinds of hierarchical models. 
For instance, a limited number of hidden units could model the set of full interactions among a small subset of visible variables, 
or, alternatively, to model all $k$-wise interactions among a large set of visible units. 
Exactly characterizing the largest hierarchical models that can be represented by an RBM is still an open problem for $n\geq 4$.

\paragraph{Universal approximation}
\label{sec:universal}

The universal approximation question asks for the smallest model within a class of models, which is able to approximate any given probability distribution on its domain to within any desired degree of accuracy. 
%
This is a special case of the problems discussed in the previous paragraphs. 
A direct consequence of Theorem~\ref{theorem:univRBMs} is 
\begin{corollary}
	\label{cor:univRBMs} 
	Let $\Xcal=\Xcal_1\times\cdots\times \Xcal_n$ and $\Ycal=\Ycal_1\times\cdots\times \Ycal_m$ be finite sets. 
	The model $\RBM_{\Xcal,\Ycal}$ is a universal approximator whenever 
	\begin{equation*}
	1+\sum_{j\in [m]}(|\Ycal_j|-1) \;\geq\; |\Xcal| / \max_{i\in [n]}|\Xcal_i|. 
	\end{equation*}
\end{corollary}
%
When all units are binary, this implies that an RBM with $2^{n-1} -1$ hidden units is a universal approximator of distributions on $\{0,1\}^n$.  
Theorem~\ref{theorem2a} improves this bound as follows: 
\begin{corollary}[{\cite[Corollary~12]{montufar2016hierarchical}}]
	\label{corollary:universal}
	Every distribution on $\{0,1\}^n$ can be approximated arbitrarily well by distributions from $\RBM_{n,m}$ whenever 
	$$m \geq \min \Big\{2^{n-1}-1, 
	\frac{2(\log(n-1)+1)}{n+1}(2^{n} -(n+1) -1) +1 \Big\}.$$ 
\end{corollary}
This is the sharpest general upper bound that is available at the moment. 
A slightly looser but simpler bound is $\frac{2(\log(n)+1)}{n+1}2^n -1$. 
Again, in specific cases there are sharper bounds available, listed in~\cite{montufar2016hierarchical}.

In terms the necessary number of hidden units for universal approximation,  bounds have been harder to obtain. 
In the general case, we only have lower bounds coming from parameter counting arguments:  
\begin{proposition}
	Let $\Mcal$ be an exponential family over $\Xcal\times \Ycal$ and $\Mcal_V$ the set of marginals on $\Xcal$. 
	If $\Mcal_V$ is a universal approximator, then $\Mcal_V$ has dimension $|\Xcal|-1$ and $\Mcal$ has dimension at least $|\Xcal|-1$. 
\end{proposition}
This implies that for $\RBM_{n,m}$ to be a universal approximator, necessarily $m\geq 2^n/(n+1) -1$. 
There is still a logarithmic gap between the upper and lower bounds. 
Further closing this gap is an important representation theoretic problem, which could help us obtain a more complete understanding of the representational power question. 
In a few small cases we can obtain the precise numbers. 
For instance, for $n=2$, the minimal size of a universal approximator is $m=1$. 
For $n=3$ it is $m=3$. 
But already for $n=4$ we can only bound the exact value between $3$ and $6$. 

\paragraph{Relative representational power}
\label{sec:relative}

As we have seen, RBMs can represent certain mixtures of product distributions. 
Complementary to this, it is natural to ask how large a mixture of products is needed in order to represent an RBM. 
%
%
Following Section~\ref{sec:definitions}, an RBM model consists of tensors which are entrywise products of tensors of with non-negative rank at most two. 
%
For many combinations of $n$ and $m$ it turns out that the RBM model represents tensors of the maximum possible rank, $2^{m}$, which implies that the smallest mixture of products that contain the RBM model is as large as one could possibly expect, having $2^m$ components: 
\begin{theorem}[{\cite[Theorem~1.2]{montufar2015does}}]
	\label{thmsummaryans}
The smallest $k$ for which the model $\Mcal_{n,k}$, consisting of arbitrary mixtures of $k$ product distributions of $n$ binary variables, contains the model $\RBM_{n,m}$, is bounded by $\frac{3}{4}n \leq \log_2(k) \leq n-1$ when $m\geq n$, by $\frac{3}{4}n \leq \log_2(k) \leq m$ when $\frac{3}{4}n\leq m\leq n$, and satisfies $\log_2(k)=m$ when $m\leq \frac{3}{4}n$. 
\end{theorem}
As shown in~\cite{montufar2015does} RBMs can express distributions with many more strong modes than mixtures of products with the same number of parameters. A strong mode is a local maximum of the probability distribution, 
with value larger than the sum of all its neighbors, whereby the vicinity structure is defined by the Hamming distance over the set of elementary events. 
Distributions with many strong modes have a large non-negative tensor rank. 
%
At the same time,~\cite{montufar2015does} shows that an RBM does not always contain a mixture of products model with the same number of parameters. 
The size of the largest mixture of products that is contained in an RBM is still an open problem. 

For hierarchical models, Lemma~\ref{generallemma} allows us to formulate an analogous result. 
The lemma implies that a hidden unit can create non-zero values of any parameter of any arbitrary hierarchical model. 
In turn, the smallest hierarchical model that contains an RBM must have all possible interactions and hence it is as large as one could possibly expect: 
\begin{proposition}
Let $n,m\in \mathbb{N}$. 
The smallest $S\subseteq 2^{V}$ for which the hierarchical model $\Ecal_S$ on $\{0,1\}^V$ contains $\RBM_{n,m}$ is $S=2^V$. 
\end{proposition}

\section{Divergence bounds}
\label{sec:approximation}

Instead of asking for the sets of distributions that can be approximated arbitrarily well by an RBM, we can take a more refined standpoint and ask for the error in the approximation of a given target distribution. 
The best possible uniform upper bound on the divergence to a model $\Mcal$ is $D_\Mcal = \max_p D(p\|\Mcal) = \max_p\inf_{q\in\Mcal}D(p\|q)$. 

Maximizing the divergence to a model, over the set of all possible targets, is an interesting problem in its own right. 
For instance, the divergence to an independence model is called multi-information and can be regarded as a measure of complexity. 
The multi-information can be used as an objective function in certain learning problems, as a way to encourage behaviors that are both predictable and diverse. 
The divergence maximization problem is challenging, even in the case of exponential families with closed formulas for the maximum likelihood estimators. 
For exponential families models the divergence maximization problem has been studied in particular by Mat\'u\v{s}~\cite{5319763}, Ay~\cite{Matus2003}, and Rauh~\cite{Rauh11:Finding_Maximizers}. 
%

In the case of RBMs, as with most machine learning models used in practice, the situation is further complicated, since we do not have closed formulas for the error minimizers of a given target. 
The approximation errors of RBMs were studied in~\cite{NIPS2011_0307} by showing that RBMs contain a number of exponential families and providing upper bounds on the divergence to such families. 
The approach was formulated more generally in~\cite{Montufar2013maximal}. 
In~\cite{montufar2014scaling} it was show how to obtain upper bounds on the expected value of the approximation error, when the target distributions are sampled from a given prior. 
In the following we discuss some of these bounds and also a divergence bound derived from the hierarchical models presented in Section~\ref{sec:representational}.

\paragraph{Upper bounds from unions of mixtures of products and hierarchical models}

The Kullback-Leibler divergence from a distribution $q$ to another distribution $p$ is 
\begin{equation*}
D(p\|q) = \sum_x p(x)\log\frac{p(x)}{q(x)}. 
\end{equation*}
Given some $p$, we are interested in the best approximation 
within a given model $\Mcal$. 
We consider the function that maps each possible target distribution $p$ to 
\begin{equation*}
D(p\|\Mcal) = \inf_{q\in\Mcal} D(p\|q). 
\end{equation*} 
The divergence to a partition model $\Pcal_A$ with blocks $A_k$, $k=1,\ldots, K$, 
is bounded above by $D(\cdot\|\Pcal_A)\leq \max_{k}\log|A_k|$. 
This bound is in fact tight. 
Corollary~\ref{lemma:distrRBMs} shows that RBMs can represent certain partition models. 
This implies the following bound. 

\begin{theorem}[{\cite[Theorem~5]{montufar2014universal}}]
	\label{theorem:univRBMs} 
	Let $\Xcal=\Xcal_1\times\cdots\times \Xcal_n$ and $\Ycal=\Ycal_1\times\cdots\times \Ycal_m$ be finite sets. 
	If $1+\sum_{j\in [m]}(|\Ycal_j|-1) \geq |\Xcal_{\Lambda\setminus\{k\}}|$ for some $\Lambda\subseteq [n]$ and $k\in\Lambda$, then 
	\begin{equation*}
	D(\cdot\|  \RBM_{\Xcal,\Ycal}) \leq \log |\Xcal_{[n]\setminus \Lambda}| . 
	\end{equation*}
\end{theorem}
Instead of partition models, we can also consider mixtures of product distributions with disjoint supports, as described in Theorem~\ref{theorem:RBMs}. 
As discussed in~\cite{NIPS2011_0307} the divergence to a mixture of models with disjoint supports can be bounded tightly from above by the maximum divergence to one of the component models over targets with the same support. 
Consider a model $\Mcal$ consisting of mixtures $\sum_j\lambda_j p_j$ of distributions $p_j\in \Mcal_j$, where $\Mcal_j$ consists of distributions supported on $A_j$, and $A_i\cap A_j=\emptyset$ whenever $i\neq j$. Then 
$$
\max_{p} D(p \|\Mcal) = \max_j\max_{p \colon \supp (p)\subseteq A_j }D(p \|\Mcal_j).
$$ 
We know that the RBM contains several mixtures of products with disjoint supports. 
Hence we can further improve the divergence upper bounds by considering the divergence to the union of all the models that are contained in the RBM model. 
This gives the following bound. 
%
\begin{theorem}[{\cite[Theorem~2]{Montufar2013maximal}}]
	\label{theorem:approximationerror}
	If $m\leq 2^{n-1}-1$, 
	\begin{equation*}
	D(\cdot \| {\RBM_{n,m}}) \leq 
	\big( n -\left\lfloor \log_2(m+1) \right\rfloor - \frac{m+1}{2^{\left\lfloor\log_2 (m+1)\right\rfloor}}  \big) \log(2) \;.\label{boundlog1}
	\end{equation*}
\end{theorem} 
A corresponding analysis for RBMs with non-binary units still needs to be worked out.

We can also bound the divergence in terms of the hierarchical models described in Theorem~\ref{theorem2a}, instead of the partition models and mixtures of products mentioned above. 
Mat\'u\v{s}~\cite{5319763} studies the divergence to hierarchical models, and proves, in particular, the following bound. 
\begin{lemma}[{\cite[Corollary~3]{5319763}}]
Consider an inclusion closed set $S\subseteq 2^V$ and the hierarchical model $\Ecal_S$ on $\{0,1\}^V$. 
Then 
$D(\cdot\|\Ecal_{S})\leq \min_{\Lambda\in S}\log|\Xcal_{V\setminus \Lambda}|$. 
\end{lemma}
In conjunction with Theorem~\ref{corollary:universal}, 
this directly implies the following bound. 
\begin{corollary}
Let $n, m\in\mathbb{N}$, and let $k$ be the largest integer with 
$m\geq \frac{\log(k)+1}{k+1}2^{k+1}-1$. 
Then $D(\cdot\| \RBM_{n,m})\leq (n-k)\log(2)$. 
\end{corollary}
A version of this result for non-binary variables and bounding the divergence to unions of hierarchical models still need to be worked out. 


\paragraph{Divergence to polyhedral exponential families}

The previous results estimate the divergence to an RBM model by looking at the divergence to exponential families or unions of exponential families that are contained within the RBM model (or within its closure, to be more precise). 
More generally, we might be interested in estimating the divergence to models whose set of log-probabilities forms a polyhedral shape, as the one shown in Figure~\ref{fig:softplus}. 
Each face of a polyhedron can be extended to an affine space, and hence corresponds to a piece of an exponential family. 	
This allows us to compute the maximum likelihood estimators of a polyhedral family in the following way. 
A related discussion was conducted recently in~\cite{SerkanPaper} in the context of mixtures of products, and in~\cite{SeigalMontufar} in the context of RBMs. 

\begin{figure}
	\centering
	\begin{tikzpicture}
	\definecolor{bd1}{rgb}{0, 0, 0.5625};
	\definecolor{bd2}{rgb}{0, 0.1250, 1.0000}
	\definecolor{bl}{rgb}{0,.06,.75};
	\tikzstyle{puntob}=[circle, draw=bl, inner sep=.05cm, fill=bl]
	
	\fill[-, color= bl,opacity=.075] (2,2) -- (0,0) -- (2,-2) -- (2,2);
	\draw (1.25,0) node {$\mathcal{M}$}; 
	\draw[-, very thick, color = bd1] (0,0) -- (2,-2);
	\draw[-, very thick, color = bd2] (2,2) -- (0,0);
	\foreach \i in {1,...,3}
	{ 
		\draw[->, color = gray] (.8*\i-2 , +2) -- (.4*\i-.1,.4*\i+.1) node [near start, above] {$ $}; %
		\draw[->, color = gray] (.8*\i-2 , -2) -- (.4*\i-.1,-.4*\i-.1) node [near start, above] {$ $}; %
	};
	\draw (-.9,1.5) node[fill=white, inner sep =.05cm] {\textcolor{bd1}{$\mathcal{M}_{0}$}}; 
	\draw (-.9,-1.5) node[fill=white, inner sep =.05cm] {\textcolor{bd2}{$\mathcal{M}_{1}$}}; 	
	\fill[-, color= bl,opacity=.0125] (-2,2) -- (0,0) -- (-2,-2) -- (-2,2);
	\foreach \i in {1,...,3}
	{ 
		\draw[->, color = gray] (-2, \i-2) -- (-.25,.1*\i-.2) node [near start, above] {$ $}; %
	};
	\draw[dashed,very thick, color = bd1] (-2,2) -- (2,-2);
	\draw[dashed,very thick, color = bd2] (-2,-2) -- (2,2);
	\draw (0,0) node[puntob] {} node[below]{\textcolor{bl}{}};		
	\end{tikzpicture}
	\caption{Illustration of the maximum likelihood projections onto a model whose log-probabilities form a polyhedron. 
		Here the polyhedron $\Mcal$ consists of the points on the positive side of two hyperplanes, $\Mcal_0$ and $\Mcal_1$. 
		Each face of the polyhedron extends to an affine space that corresponds to an exponential family. 
		For each possible target, each exponential family has a unique maximum likelihood projection point. 
		Arrows indicate how targets project to the different faces of $\Mcal$. }
	\label{fig:projections}
\end{figure}
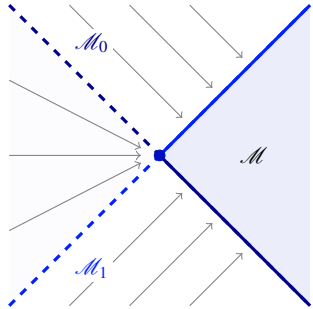

Given a target distribution $p$ and a model with log-probabilities from a polyhedron $\Mcal$ we proceed as follows. 
\begin{itemize}
\item 
For each face $\Mcal_i$ of $\Mcal$, we define a corresponding exponential family $\Ecal_i$. 
Any basis of the affine hull of $\Mcal_i$ forms a sufficient statistics, and we can take any point in $\Mcal_i$ as a reference measure. 
\item Then we compute the maximum likelihood estimator $q_i = \operatorname{arginf}_{q\in\Ecal_i} D(p\| q )$ for each individual exponential family $\Ecal_i$. 
For exponential families the maximum likelihood estimation problem is concave and has a unique solution (possibly on the closure of the exponential family). 
\item 
Then we verify which of the projections $q_i$ are feasible, meaning that they satisfy the constraints of the corresponding face $\Mcal_i$. 
\item 
Finally, we select among the feasible projections, the one with the smallest divergence to the target distribution $p$. 
This is illustrated in Figure~\ref{fig:projections}. 
\end{itemize}

\paragraph{Tightness of the bounds}

In the previous paragraphs we provided upper bounds on the divergence from arbitrary target distributions to an RBM model. 
One may wonder about the tightness of these bounds. 
For the special case of independence models, which are RBMs with no hidden units, the bounds are tight, provided all visible variables have state spaces of equal cardinality. 
However, already in the case of one single hidden unit, the exact value of the maximum divergence is not known in general. 

Experiments on small RBMs~\cite{NIPS2011_0307,montufar2014universal} seem to indicate that the bounds provided in the previous paragraphs are in good agreement with the actual values. 
Empirical studies are difficult because of two opposing effects. 
On the one hand, sequential optimization methods may only lead to sub-optimal approximations of a given target. 
In fact, part of the motivation for deriving theoretical upper bounds is to monitor the quality of our sequential optimization methods. 
On the other hand, finding a target distribution with maximum divergence to the model may be a difficult problem itself. It may be that the vast majority of possible targets are not as far to the model as the divergence maximizer. In turn, the theoretical upper bounds could appear pessimistic for most of the targets. 
In~\cite{montufar2014scaling} it is shown how to estimate the expected value of the divergence when the target distributions are sampled from a Dirichlet distribution. The average values tend to be indeed much lower than the maximum values. 

A recent work~\cite{SeigalMontufar} shows that the model $\RBM_{3,2}$ has a boundary described in terms of a union of exponential families, and uses this description to obtain the divergence maximizers to the model. 
It shows that the divergence bounds obtained in Theorem~\ref{theorem:approximationerror} are tight for this particular model. 


\begin{theorem}[{\cite[Theorem~3]{SeigalMontufar}}]
	\label{proposition:divergencemaximizers}
	The maximum divergence to $\RBM_{3,2}$ is $\frac{1}{2}\log 2$. 
	The maximizers are 
	$\frac{1}{4}(\delta_{000} + \delta_{011} + \delta_{101} + \delta_{110})$ and 
	$\frac{1}{4}(\delta_{001} + \delta_{010} + \delta_{100} + \delta_{111})$. 
For each of these targets, there is one distinct projection point on each of the six boundary pieces of $\RBM_{3,2}$. 
\end{theorem}


\section{Implicit description}
\label{sec:semialgebraic}

So far we have discussed probability models presented explicitly, as parametric families of distributions. 
RBMs can also be expressed implicitly, in terms of constraints that apply to the distributions within the model, and only to the distributions within the model. 
Indeed, since RBMs have a polynomial parametrization, they can be described semi-algebraically as the set of real solutions to a collection of polynomial equations and polynomial inequalities. 
The \emph{implicitization problem} consists of replacing a parametric description with a description as the solution set of a collection of equations and inequalities. 
Finding implicit characterizations for graphical models with hidden variables is a significant challenge and a central topic within algebraic statistics~\cite{drton2009lectures,sullivant2018}. 
In principle both, explicit and implicit presentations, can be challenging to interpret in general, for instance when the parametrization is convoluted, or when the constraints correspond to complicated properties of the distributions. 
However, in some cases the implicit descriptions have a very intuitive statistical interpretation and can allow us to make significant advances over what is possible with a parametric description alone. 
Implicit descriptions can be extremely useful for hypothesis testing, membership testing, and other related problems. 
So far there are not many results on the implicit description of RBMs. 
The following discussion is intended as a motivation.

\paragraph{Markov properties}

A fully observable undirected graphical model can be defined in terms of the factorization property~\eqref{eq:factorizationhierarchical}. 
Each of the factors can be considered as a parameter, 
or can be easily parametrized, as shown in~\eqref{eq:hierarchical}. 
%
%
Graphical models are usually also motivated and defined in terms of so-called Markov properties, 
or conditional independence statements. 
These are constraints that characterize the probability distributions in the model. 
Undirected graphical models encode conditional independence relations in terms of the structure of the graph. 
Specifically, a probability distribution is contained in an undirected graphical model with graph $G$ if and only if it satisfies all conditional independence statements encoded by the graph $G$, namely 
\begin{equation}
X_A \independent X_B \; |\; X_C, 
\label{eq:ci}
\end{equation}
whenever $A,B,C$ are disjoint subsets of $V$ for which any path connecting a point in $A$ and a point in $B$, passes through $C$. 
Equation~\eqref{eq:ci} means that $p$ satisfies the equations 
$p(x_A,x_B|x_C) = p(x_A|x_C) p(x_B|x_C)$, or, equivalently, 
\begin{equation*}
p(x_A,x_B,x_C)\sum_{x_A',x_B'}p(x_A',x_B',x_C) - \sum_{x_B'} p(x_A,x_B',x_C) \sum_{x_A'} p(x_A',x_B,x_C) = 0, 
\end{equation*}
for all  $x_A\in\Xcal_A$, $x_B\in\Xcal_B$, $x_C\in \Xcal_C$. 
These are quadratic binomial equations in the indeterminates $p(x)\in\mathbb{R}$, $x\in\Xcal$. 
A famous theorem by Hammersley and Clifford~\cite{HammersleyClifford:1971} gives the correspondence between the conditional independence constraints and the factorization property of the joint distributions in a fully observable graphical model. 
This correspondence is usually limited to strictly positive probability distributions. 
For distributions that are not strictly positive, which lie at the boundary of the probability simplex, the correspondence is more subtle in general and has been investigated in~\cite{geiger2006}. 
%
The main point here is that we can formulate a parametric set of functions in terms of constraints, or properties of distributions. 
Moreover, at least in the case of fully observable undirected graphical models, the constraints have an intuitive statistical interpretation.

\paragraph{Constraints in a small RBM}

A natural question is what are the constraints that define the visible distributions in a an RBM, and more generally, in a hierarchical model with hidden variables. 
Aside from RBMs with one single hidden unit, which correspond to mixtures of two product distributions, 
the RBM with 4 visible and 2 hidden variables has been studied, which turns out to be a hyper-surface defined as the zero set of a polynomial with over a trillion monomials~\cite{Cueto:2010:ICB:1866469.1866627}. 

The constraints that apply to $\RBM_{3,2}$ were studied in~\cite{montufar2015does}, obtaining a coarse description of the model. 
The full semi-algebraic description of this model was then obtained in~\cite{SeigalMontufar}. 
The characterization is as follows. 

\begin{theorem}[{\cite[Theorem~1]{SeigalMontufar}}] \label{prop}
	The model $\RBM_{3,2}$ is the union of six basic semi-algebraic sets, each described by two inequalities, namely: 
	$$ \begin{matrix} 
	\{ p_{000} p_{011} \geq p_{001} p_{010} , \quad p_{100} p_{111} \geq p_{101} p_{110} \}\phantom{.}\\ 
	\{ p_{000} p_{011} \leq p_{001} p_{010} , \quad p_{100} p_{111} \leq p_{101} p_{110} \} \phantom{.}\\ 
	\{ p_{000} p_{101} \geq p_{001} p_{100} , \quad p_{010} p_{111} \geq p_{011} p_{110} \} \phantom{.}\\
	\{ p_{000} p_{101} \leq p_{001} p_{100} , \quad p_{010} p_{111} \leq p_{011} p_{110} \} \phantom{.}\\
	\{ p_{000} p_{110} \geq p_{100} p_{010} , \quad p_{001} p_{111} \geq p_{101} p_{011} \} \phantom{.} \\
	\{ p_{000} p_{110} \leq p_{100} p_{010} , \quad p_{001} p_{111} \leq p_{101} p_{011} \} . \end{matrix} $$
\end{theorem}
Each pair of inequalities represents the non-negativity or non-positivity of two determinants. 
These determinants capture the conditional correlations of two of the variables, given the value of the third variable. 
The conditional correlation is either non-negative or non-positive for both possible values of the third variable. 

This theorem gives a precise description of the geometry of the model. 
The model is full dimensional in the ambient probability simplex. 
Hence the description involves only inequalities and no equations (aside from the normalization constraint $\sum_x p_x=1$). 
%
Setting either of the inequalities to an equation gives a piece of the boundary of the model. 
Each boundary piece is an exponential family which can be interpreted as the set of mixtures of one arbitrary product distribution and one product distribution with support on the states with fixed value of one of the variables, similar to the distributions described in Theorem~\ref{theorem:RBMs}. 
For these exponential families we can compute the maximum likelihood estimators in closed form, as described in the previous paragraph, and 
also obtain the exact maximizers of the divergence, given in Theorem~\ref{proposition:divergencemaximizers}. 
With the implicit description at hand~\cite{SeigalMontufar} also shows that the model $\RBM_{3,2}$ is equal to the mixture model of three product distributions, and that it does not contain any distributions with $4$ modes, both statements that had been conjectured in~\cite{montufar2015does}. 

\paragraph{Coarse necessary constraints}

Obtaining the exact constraints that define an RBM model can be difficult in general. 
In Section~\ref{sec:representational} we described submodels of the RBM, which can be interpreted as constraints that are sufficient for probability distributions to be contained in the model, but not necessary. 
A complementary alternative is to look for constraints that are necessary for distributions to be in the model, but not sufficient. 
These sometimes are easier to obtain and interpret. 
An example are strong mode inequalities in mixtures of product distributions~\cite{montufar2015does}, 
and information theoretic inequalities in Bayesian networks~\cite{e17042304}. 
Mode inequality constraints for RBMs have been studied in~\cite{montufar2015does}. 
Another possible direction was suggested in~\cite{SeigalMontufar}, namely to consider the inequality constraints that apply to mixtures of two product distributions and how they combine when building Hadamard products.

\section{Open problems}
\label{sec:outlook}

The theory of RBMs is by no means a finished subject. 
In the following, I collect a selection of problems, as a sort of work program, addressing which I think is important towards obtaining a more complete picture of RBMs and advancing the theory of graphical models with hidden variables in general. 

\begin{enumerate}
	\item 
Can we find non-trivial constraints on the sets of representable probability distributions? 
A related type of questions has been investigated in~\cite{montufar2015does}, with focus on the approximation of distributions with many modes, or mixtures of product distributions. 
	\item Closely related to the previous item, given the number $n$ of visible units, what is the smallest number $m$ of hidden units for which $\RBM_{n,m}$ is a universal approximator? 
Alternatively, can we obtain lower bounds on the number of hidden units of an RBM that is a universal approximator? Here, of course, we are interested in lower bounds that do not readily follow from parameter counting arguments. 	
	The first open case is $n=4$, for which we have bounds $3\leq m\leq 6$. 
	\item What is the smallest tropical RBM that is a universal approximator? 
	Equivalently, what is the smallest $m$ for which a sum of one affine function and $m$ ReLUs can express any function of $n$ binary variables?  
	\item Characterize the support sets of the distributions in the closure of an RBM. 
	We note that characterizing the support sets of distributions in the closure of an exponential family corresponds to describing the faces its convex support polytope. 
	\item Also in relation to the first item, obtain an implicit description of the RBM model.  
	The work~\cite{SeigalMontufar} gives the description of $\RBM_{3,2}$ and ideas for the inequality constraints of larger models. 
	Interesting cases to consider are $\RBM_{4,3}$ (this might be the full probability simplex), $\RBM_{5,2}$, $\RBM_{6,5}$. 
	For the latter~\cite{montufar2015does} obtained some linear inequality constraints. 
	\item Can we produce explicit descriptions of the maximum likelihood estimators? 
	Here~\cite{SeigalMontufar} indicates possible avenues. 
	\item Describe the structure of the likelihood function of an RBM. 
	In particular, what is the number of local and global optimizers? 
	How does this number depend on the empirical data distribution? 
	\item Describe the critical points of the EM algorithm for an RBM model or for its Zariski closure. 
	\item Characterize the sets of parameters that give rise to the different distributions expressible by an RBM. 
	When this is finite, are there parameter symmetries other than those coming from relabeling units and states? 
	\item What is the maximum possible value of the divergence to an RBM model, 
	$D_{n,m} = \max_{p\in\Delta_{\{0,1\}^n}}\inf_{q\in\RBM_{n,m}} D(p\|q)$, 
	and what are the divergence maximizers?  
	We know $D_{3,0}=2 \log 2$ from results for independence models (see, e.g.,~\cite{Montufar2013maximal}), and $D_{3,2}=\frac{1}{2}\log 2$ (see Theorem~\ref{prop} and ~\cite{SeigalMontufar}). 
The first open case is $D_{3,1}$. Discussions with Johannes Rauh suggest $-\frac{3}{4}\log_2(2\sqrt{3}-3)$. 
\item In relation to the previous item, can we provide lower bounds on the maximum divergence from a given union of exponential families?  
	\item Does the tropical RBM model have the expected dimension? 
	In~\cite{Cueto2010} it was conjectured that it does. 
	The problem remains open, even though~\cite{montufar2015dimension} gave a proof for the RBM. The description of the tropical RBM as a superposition of ReLUs could be useful here. 
	\item What is the largest mixture of product distributions that is contained in the RBM model? 
	A result from~\cite{montufar2015does} shows that RBMs do not always contain mixtures of products of the same dimension. 
	\item What are the largest hierarchical models that are contained in the closure of an RBM model? 
	A partial characterization of the polynomials that are expressible in terms of soft-plus and rectified linear units on binary inputs was obtained in~\cite{montufar2016hierarchical}. A full characterization is still missing.  
	\item Generalize the analysis of hierarchical models contained in RBM models to the case of non-binary variables (both visible and hidden).  
\end{enumerate}

%

\begin{acknowledgement}	
I thank Shun-ichi Amari for inspiring discussions over the years. 
This review article originated at the IGAIA IV conference in 2016 dedicated to his 80th birthday. 
I am grateful to Nihat Ay, Johannes Rauh, Jason Morton, and more recently Anna Seigal for our collaborations. 
I thank Fero Mat\'u\v{s} for discussions on the divergence maximization for hierarchical models, lastly at the MFO Algebraic Statistics meeting in 2017. 
I thank Bernd Sturmfels for many fruitful discussions, and Dave Ackley for insightful discussions at the Santa Fe Institute in 2016. 	
\end{acknowledgement}

\bibliography{referenzen}

\begin{thebibliography}{10}

\bibitem{Ackley85alearning}
D.~H. Ackley, G.~E. Hinton, and T.~J. Sejnowski.
\newblock A learning algorithm for {B}oltzmann machines.
\newblock {\em Cognitive Science}, pages 147--169, 1985.

\bibitem{SerkanPaper}
E.~Allman, H.~B. Cervantes, R.~Evans, S.~Ho\c{s}ten, K.~Kubjas, D.~Lemke,
  J.~Rhodes, and P.~Zwiernik.
\newblock Maximum likelihood estimation of the latent class model through model
  boundary decomposition.
\newblock 2017.

\bibitem{amari1985differential}
S.~Amari.
\newblock {\em Differential-geometrical methods in statistics}.
\newblock Lecture notes in statistics. Springer-Verlag, 1985.

\bibitem{Amari:1998:NGW:287476.287477}
S.~Amari.
\newblock Natural gradient works efficiently in learning.
\newblock {\em Neural Comput.}, 10(2):251--276, Feb. 1998.

\bibitem{Amari99informationgeometry}
S.~Amari.
\newblock Information geometry on hierarchical decomposition of stochastic
  interactions.
\newblock {\em IEEE Transaction on Information Theory}, 47:1701--1711, 1999.

\bibitem{igaia}
S.~Amari.
\newblock {\em Information Geometry and its Applications}, volume 194 of {\em
  Applied Mathematical Sciences}.
\newblock Springer Japan, 2016.

\bibitem{125867}
S.~Amari, K.~Kurata, and H.~Nagaoka.
\newblock Information geometry of {B}oltzmann machines.
\newblock {\em IEEE Transactions on Neural Networks}, 3(2):260--271, Mar 1992.

\bibitem{amari2007methods}
S.~Amari and H.~Nagaoka.
\newblock {\em Methods of Information Geometry}.
\newblock Translations of mathematical monographs. American Mathematical
  Society, 2007.

\bibitem{PhysRevX.8.021050}
M.~H. Amin, E.~Andriyash, J.~Rolfe, B.~Kulchytskyy, and R.~Melko.
\newblock Quantum {B}oltzmann machine.
\newblock {\em Phys. Rev. X}, 8:021050, May 2018.

\bibitem{aoyagi:2010}
M.~Aoyagi.
\newblock Stochastic complexity and generalization error of a {R}estricted
  {B}oltzmann {M}achine in {B}ayesian estimation.
\newblock {\em Journal of Machine Learning Research}, 99:1243--1272, August
  2010.

\bibitem{InfGeo}
N.~Ay, J.~Jost, H.~L\^{e}, and L.~Schwachh\"ofer.
\newblock {\em Information Geometry}, volume~64 of {\em Ergebnisse der
  Mathematik und ihrer Grenzgebiete}.
\newblock Springer, 2017.

\bibitem{Bengio-2009}
Y.~Bengio.
\newblock Learning deep architectures for {AI}.
\newblock {\em Foundations and Trends in Machine Learning}, 2(1):1--127, 2009.
\newblock Also published as a book. Now Publishers, 2009.

\bibitem{Bieri1984}
R.~Bieri and J.~Groves.
\newblock The geometry of the set of characters iduced by valuations.
\newblock {\em Journal f\"ur die reine und angewandte Mathematik},
  347:168--195, 1984.

\bibitem{Brown86:Fundamentals_of_Exponential_Families}
L.~Brown.
\newblock {\em Fundamentals of Statistical Exponential Families: With
  Applications in Statistical Decision Theory}.
\newblock Institute of Mathematical Statistics, Hayworth, CA, USA, 1986.

\bibitem{Catalisano2011}
M.~Catalisano, A.~Geramita, and A.~Gimigliano.
\newblock Secant varieties of $\mathbb{P}^1\times\dots\times\mathbb{P}^1$
  ($n$-times) are not defective for $n\geq5$.
\newblock {\em Journal of Algebraic Geometry}, 20:295--327, 2011.

\bibitem{csiszar_information_1984}
I.~Csisz\'{a}r and G.~Tusn\'{a}dy.
\newblock {Information Geometry and Alternating minimization procedures}.
\newblock {\em Statistics and Decisions}, Supplement Issue 1, 1984.

\bibitem{Cueto2010}
M.~A. Cueto, J.~Morton, and B.~Sturmfels.
\newblock Geometry of the restricted {B}oltzmann machine.
\newblock In M.~A.~G. Viana and H.~P. Wynn, editors, {\em Algebraic methods in
  statistics and probability II, AMS Special Session}, volume~2. American
  Mathematical Society, 2010.

\bibitem{Cueto:2010:ICB:1866469.1866627}
M.~A. Cueto, E.~A. Tobis, and J.~Yu.
\newblock An implicitization challenge for binary factor analysis.
\newblock {\em Journal of Symbolic Computation}, 45(12):1296--1315, 2010.

\bibitem{10.2307/2984875}
A.~P. Dempster, N.~M. Laird, and D.~B. Rubin.
\newblock Maximum likelihood from incomplete data via the em algorithm.
\newblock {\em Journal of the Royal Statistical Society. Series B
  (Methodological)}, 39(1):1--38, 1977.

\bibitem{Draisma}
J.~Draisma.
\newblock A tropical approach to secant dimensions.
\newblock {\em J. Pure Appl. Algebra}, 212(2):349--363, 2008.

\bibitem{drton2009lectures}
M.~Drton, B.~Sturmfels, and S.~Sullivant.
\newblock {\em Lectures on Algebraic Statistics}.
\newblock Oberwolfach Seminars. Springer Verlag, 2009.

\bibitem{fischer2009contrastive}
A.~Fischer and C.~Igel.
\newblock Contrastive divergence learning may diverge when training restricted
  {B}oltzmann machines.
\newblock In {\em Frontiers in Computational Neuroscience. Bernstein Conference
  on Computational Neuroscience (BCCN 2009)}, 2009.

\bibitem{fischer2010bounding}
A.~Fischer and C.~Igel.
\newblock Bounding the bias of contrastive divergence learning.
\newblock {\em Neural Computation}, 23(3):664--673, 2010.

\bibitem{10.1007/978-3-642-33275-3_2}
A.~Fischer and C.~Igel.
\newblock An introduction to restricted {B}oltzmann machines.
\newblock In L.~Alvarez, M.~Mejail, L.~Gomez, and J.~Jacobo, editors, {\em
  Progress in Pattern Recognition, Image Analysis, Computer Vision, and
  Applications}, pages 14--36, Berlin, Heidelberg, 2012. Springer Berlin
  Heidelberg.

\bibitem{fischer2014training}
A.~Fischer and C.~Igel.
\newblock Training restricted {B}oltzmann machines: an introduction.
\newblock {\em Pattern Recognition}, 47(1):25--39, 2014.

\bibitem{Fischer15PTBound}
A.~Fischer and C.~Igel.
\newblock A bound for the convergence rate of parallel tempering for sampling
  restricted {B}oltzmann machines.
\newblock {\em Theoretical Computer Science}, 598:102 -- 117, 2015.

\bibitem{NIPS1991_535}
Y.~Freund and D.~Haussler.
\newblock Unsupervised learning of distributions on binary vectors using two
  layer networks.
\newblock In J.~E. Moody, S.~J. Hanson, and R.~P. Lippmann, editors, {\em
  Advances in Neural Information Processing Systems 4}, pages 912--919.
  Morgan-Kaufmann, 1992.

\bibitem{Freund:1994:ULD:902676}
Y.~Freund and D.~Haussler.
\newblock Unsupervised learning of distributions on binary vectors using two
  layer networks.
\newblock Technical report, Santa Cruz, CA, USA, 1994.

\bibitem{geiger2006}
D.~Geiger, C.~Meek, and B.~Sturmfels.
\newblock On the toric algebra of graphical models.
\newblock {\em Ann. Statist.}, 34(3):1463--1492, 06 2006.

\bibitem{gibbs1902elementary}
J.~Gibbs.
\newblock {\em Elementary Principles in Statistical Mechanics: Developed with
  Especial Reference to the Rational Foundations of Thermodynamics}.
\newblock Elementary Principles in Statistical Mechanics: Developed with
  Especial Reference to the Rational Foundation of Thermodynamics. C.
  Scribner's sons, 1902.

\bibitem{pmlr-v37-grosse15}
R.~Grosse and R.~Salakhudinov.
\newblock Scaling up natural gradient by sparsely factorizing the inverse
  fisher matrix.
\newblock In F.~Bach and D.~Blei, editors, {\em Proceedings of the 32nd
  International Conference on Machine Learning}, volume~37 of {\em Proceedings
  of Machine Learning Research}, pages 2304--2313, Lille, France, 07--09 Jul
  2015. PMLR.

\bibitem{HammersleyClifford:1971}
J.~M. Hammersley and P.~E. Clifford.
\newblock Markov random fields on finite graphs and lattices.
\newblock {\em {\em Unpublished manuscript}}, 1971.

\bibitem{Hinton2002}
G.~E. Hinton.
\newblock Training products of experts by minimizing contrastive divergence.
\newblock {\em Neural Computation}, 14:1771--1800, 2002.

\bibitem{Hinton2010}
G.~E. Hinton.
\newblock A practical guide to training restricted {B}oltzmann machines,
  version 1.
\newblock Technical report, UTML2010-003, University of Toronto, 2010.

\bibitem{Hinton:2006:FLA:1161603.1161605}
G.~E. Hinton, S.~Osindero, and Y.-W. Teh.
\newblock A fast learning algorithm for deep belief nets.
\newblock {\em Neural Computation}, 18(7):1527--1554, July 2006.

\bibitem{hinton1983}
G.~E. Hinton and T.~J. Sejnowski.
\newblock Analyzing cooperative computation.
\newblock In {\em Proceedings of the Fifth Annual Conference of the Cognitive
  Science Society, Rochester NY}, 1983.

\bibitem{Hinton:1986:LRB:104279.104291}
G.~E. Hinton and T.~J. Sejnowski.
\newblock Parallel distributed processing: Explorations in the microstructure
  of cognition, vol. 1.
\newblock chapter Learning and Relearning in Boltzmann Machines, pages
  282--317. MIT Press, Cambridge, MA, USA, 1986.

\bibitem{Hopfield:1988:NNP:65669.104422}
J.~J. Hopfield.
\newblock Neurocomputing: Foundations of research.
\newblock chapter Neural Networks and Physical Systems with Emergent Collective
  Computational Abilities, pages 457--464. MIT Press, Cambridge, MA, USA, 1988.

\bibitem{huang2000statistical}
K.~Huang.
\newblock {\em Statistical Mechanics}.
\newblock John Wiley and Sons, 2000.

\bibitem{jordan2004}
M.~I. Jordan.
\newblock Graphical models.
\newblock {\em Statist. Sci.}, 19(1):140--155, 02 2004.

\bibitem{Karakida:2016:DAC:2949079.2949224}
R.~Karakida, M.~Okada, and S.~Amari.
\newblock Dynamical analysis of contrastive divergence learning: Restricted
  {B}oltzmann machines with {G}aussian visible units.
\newblock {\em Neural Networks}, 79:78--87, July 2016.

\bibitem{lauritzen1996}
S.~L. Lauritzen.
\newblock {\em Graphical Models}.
\newblock Oxford University Press, 1996.

\bibitem{LeRoux:2008:RPR:1374176.1374187}
N.~Le~Roux and Y.~Bengio.
\newblock Representational power of restricted {B}oltzmann machines and deep
  belief networks.
\newblock {\em Neural Computation}, 20(6):1631--1649, June 2008.

\bibitem{wli}
W.~Li and G.~Mont\'{u}far.
\newblock Natural gradient via optimal transport {I}.
\newblock {\em arXiv preprint arXiv:1803.07033}, 2018.

\bibitem{NIPS2013_5020}
J.~Martens, A.~Chattopadhya, T.~Pitassi, and R.~Zemel.
\newblock On the representational efficiency of restricted {B}oltzmann
  machines.
\newblock In {\em Advances in Neural Information Processing Systems 26}, pages
  2877--2885. Curran Associates, Inc., 2013.

\bibitem{Matus2003}
F.~Mat{\'u}{\v{s}} and N.~Ay.
\newblock On maximization of the information divergence from an exponential
  family.
\newblock In {\em Proceedings of the {WUPES}'03}, pages 199--204, 2003.

\bibitem{5319763}
F.~Mat\'u\v{s}.
\newblock Divergence from factorizable distributions and matroid
  representations by partitions.
\newblock {\em Information Theory, IEEE Transactions on}, 55(12):5375--5381,
  Dec 2009.

\bibitem{Montavon:2016:WTR:3157382.3157513}
G.~Montavon, K.-R. M\"{u}ller, and M.~Cuturi.
\newblock Wasserstein training of restricted boltzmann machines.
\newblock In {\em Proceedings of the 30th International Conference on Neural
  Information Processing Systems}, NIPS'16, pages 3718--3726, USA, 2016. Curran
  Associates Inc.

\bibitem{montufar2014universal}
G.~Mont{\'u}far.
\newblock Universal approximation depth and errors of narrow belief networks
  with discrete units.
\newblock {\em Neural Computation}, 26(7):1386--1407, 2014.

\bibitem{montufar2014deep}
G.~Mont{\'u}far.
\newblock Deep narrow {B}oltzmann machines are universal approximators.
\newblock In {\em International Conference on Learning Representations (ICLR
  15)}, 2015.
\newblock Published online at arXiv:1411.3784.

\bibitem{montufar2011refinements}
G.~Mont\'ufar and N.~Ay.
\newblock Refinements of universal approximation results for deep belief
  networks and restricted {B}oltzmann machines.
\newblock {\em Neural Computation}, 23(5):1306--1319, 2011.

\bibitem{montufar2014expressive}
G.~Mont{{\'u}}far, N.~Ay, and K.~Ghazi-Zahedi.
\newblock Geometry and expressive power of conditional restricted {B}oltzmann
  machines.
\newblock {\em Journal of Machine Learning Research}, 16:2405--2436, 2015.

\bibitem{montufar2013discrete}
G.~Mont\'ufar and J.~Morton.
\newblock Discrete restricted {B}oltzmann machines.
\newblock In {\em Online Proceedings of the 1-st International Conference on
  Learning Representations (ICLR2013)}, 2013.

\bibitem{montufar2015does}
G.~Mont\'ufar and J.~Morton.
\newblock When does a mixture of products contain a product of mixtures?
\newblock {\em SIAM Journal on Discrete Mathematics}, 29(1):321--347, 2015.

\bibitem{montufar2015dimension}
G.~Mont\'ufar and J.~Morton.
\newblock Dimension of marginals of {K}ronecker product models.
\newblock {\em SIAM Journal on Applied Algebra and Geometry}, 1(1):126--151,
  2017.

\bibitem{montufar2014scaling}
G.~Mont{\'u}far and J.~Rauh.
\newblock Scaling of model approximation errors and expected entropy distances.
\newblock {\em Kybernetika}, 50(2):234--245, 2014.

\bibitem{montufar2016hierarchical}
G.~Mont\'ufar and J.~Rauh.
\newblock Hierarchical models as marginals of hierarchical models.
\newblock {\em International Journal of Approximate Reasoning}, 88(Supplement
  C):531--546, 2017.

\bibitem{NIPS2011_0307}
G.~Mont\'ufar, J.~Rauh, and N.~Ay.
\newblock Expressive power and approximation errors of restricted {B}oltzmann
  machines.
\newblock In {\em Advances in Neural Information Processing Systems 24}, pages
  415--423, 2011.

\bibitem{Montufar2013maximal}
G.~Mont{\'u}far, J.~Rauh, and N.~Ay.
\newblock {\em Geometric Science of Information: First International
  Conference, GSI 2013, Paris, France, August 28-30, 2013. Proceedings},
  chapter Maximal Information Divergence from Statistical Models Defined by
  Neural Networks, pages 759--766.
\newblock Springer Berlin Heidelberg, Berlin, Heidelberg, 2013.

\bibitem{Pascanu+Bengio-ICLR2014}
R.~Pascanu and Y.~Bengio.
\newblock Revisiting natural gradient for deep networks.
\newblock In {\em International Conference on Learning Representations 2014
  (Conference Track)}, Apr. 2014.

\bibitem{Rao45statmanifold}
R.~C. Rao.
\newblock {Information and the accuracy attainable in the estimation of
  statistical parameters}.
\newblock {\em Bull. Calcutta Math. Soc.}, 37:81--91, 1945.

\bibitem{Rauh11:Finding_Maximizers}
J.~Rauh.
\newblock Finding the maximizers of the information divergence from an
  exponential family.
\newblock {\em IEEE Transactions on Information Theory}, 57(6):3236--3247,
  2011.

\bibitem{Rauh13:Optimal_Expfams}
J.~Rauh.
\newblock Optimally approximating exponential families.
\newblock {\em Kybernetika}, 49(2):199--215, 2013.

\bibitem{Salakhutdinov08learningand}
R.~Salakhutdinov.
\newblock Learning and evaluating {B}oltzmann machines.
\newblock Technical report, 2008.

\bibitem{NIPS2009_3717}
R.~Salakhutdinov.
\newblock Learning in {M}arkov random fields using tempered transitions.
\newblock In Y.~Bengio, D.~Schuurmans, J.~D. Lafferty, C.~K.~I. Williams, and
  A.~Culotta, editors, {\em Advances in Neural Information Processing Systems
  22}, pages 1598--1606. Curran Associates, Inc., 2009.

\bibitem{SalHinton07}
R.~Salakhutdinov and G.~E. Hinton.
\newblock Deep {B}oltzmann machines.
\newblock In {\em Proceedings of the International Conference on Artificial
  Intelligence and Statistics (AISTATS 09)}, pages 448--455, 2009.

\bibitem{Salakhutdinov:2007}
R.~Salakhutdinov, A.~Mnih, and G.~E. Hinton.
\newblock Restricted {B}oltzmann machines for collaborative filtering.
\newblock In {\em Proceedings of the 24th international conference on Machine
  learning}, ICML~'07, pages 791--798, New York, NY, USA, 2007. ACM.

\bibitem{SeigalMontufar}
A.~Seigal and G.~Mont{\'u}far.
\newblock Mixtures and products in two graphical models.
\newblock {\em To appear in Journal of Algebraic Statistics}, 2018.
\newblock Preprint available from arXiv:1709.05276.

\bibitem{Sejnowski86higher-orderboltzmann}
T.~J. Sejnowski.
\newblock Higher-order {B}oltzmann machines.
\newblock In {\em Neural Networks for Computing}, pages 398--403. American
  Institute of Physics, 1986.

\bibitem{Smolensky:1986:IPD:104279.104290}
P.~Smolensky.
\newblock Parallel distributed processing: Explorations in the microstructure
  of cognition, vol. 1.
\newblock chapter Information Processing in Dynamical Systems: Foundations of
  Harmony Theory, pages 194--281. MIT Press, Cambridge, MA, USA, 1986.

\bibitem{e17042304}
B.~Steudel and N.~Ay.
\newblock Information-theoretic inference of common ancestors.
\newblock {\em Entropy}, 17(4):2304, 2015.

\bibitem{sullivant2018}
S.~Sullivant.
\newblock {\em Algebraic Statistics}.
\newblock 2018.

\bibitem{DBLP:journals/neco/SutskeverH08}
I.~Sutskever and G.~E. Hinton.
\newblock Deep, narrow sigmoid belief networks are universal approximators.
\newblock {\em Neural Computation}, 20(11):2629--2636, 2008.

\bibitem{Tieleman:2008:TRB:1390156.1390290}
T.~Tieleman.
\newblock Training restricted {B}oltzmann machines using approximations to the
  likelihood gradient.
\newblock In {\em Proceedings of the 25th International Conference on Machine
  Learning}, ICML~'08, pages 1064--1071, New York, NY, USA, 2008. ACM.

\bibitem{Watanabe:2009:AGS:1655832}
S.~Watanabe.
\newblock {\em Algebraic Geometry and Statistical Learning Theory}.
\newblock Cambridge University Press, New York, NY, USA, 2009.

\bibitem{welling:exponential}
M.~{Welling}, M.~{Rosen-Zvi}, and G.~E. {Hinton}.
\newblock Exponential family harmoniums with an application to information
  retrieval.
\newblock In {\em Advances in Neural Information Processing Systems 17}, pages
  1481--1488, 2005.

\bibitem{Younes1996109}
L.~Younes.
\newblock Synchronous {B}oltzmann machines can be universal approximators.
\newblock {\em Applied Mathematics Letters}, 9(3):109--113, 1996.

\bibitem{ZHANG20181186}
N.~Zhang, S.~Ding, J.~Zhang, and Y.~Xue.
\newblock An overview on restricted {B}oltzmann machines.
\newblock {\em Neurocomputing}, 275:1186 -- 1199, 2018.

\end{thebibliography}
\bibliographystyle{abbrv}

\end{document}